\pdfoutput=1
\documentclass{article}
\usepackage[final]{corl_2026}
\makeatletter
\renewcommand{\@conferencelocation}{Austin, Texas, USA}
\makeatother
\usepackage{amsmath, amssymb}
\usepackage{multirow}
\usepackage{booktabs}
\usepackage{graphicx}

\hypersetup{
  pdftitle={What Makes Video World Model Latents Action-Relevant: Prediction over Reconstruction},
  pdfauthor={Jewon Yeom, Hanseul Kim, Jeongjae Park, Sungmok Jung, Jaejin Lee, Taesup Kim}
}

\title{What Makes Video World Model Latents Action-Relevant: Prediction over Reconstruction}

\author{
  Jewon Yeom\thanks{Equally Contributed. \texttt{\{jewon0908, k1seul\}@snu.ac.kr}} \enspace
  Hanseul Kim\footnotemark[1] \enspace
  Jeongjae Park \enspace
  Sungmok Jung \enspace
  Jaejin Lee \enspace
  Taesup Kim\thanks{Corresponding Author. \texttt{taesup.kim@snu.ac.kr}} \\
  Graduate School of Data Science, Seoul National University
}

\begin{document}
\maketitle

\begin{abstract}
Video world models increasingly provide visual representations for robot policies, yet the relation between their pretraining signals and action-relevant latent structure remains unclear. We compare image-only, video-pretrained, reconstruction-based, diffusion, and dynamics encoders using a common inverse-dynamics action probe. On a fixed four-task held-out subset of LIBERO, high pixel fidelity does not imply action recoverability, and video-pretrained encoders yield the strongest action readouts. Matched fine-tuning controls evaluated over all $26$ held-out tasks show that correct transition--action pairing, rather than adaptation alone or generic next-state prediction, drives the large relative gains for V-JEPA~2 and VideoMAE; Web-DINO gains much less. On the fixed subset, inverse-dynamics supervision also improves robustness to visual corruption, while CALVIN shows that fixed environments can narrow the gap between video and image encoders. In a limited closed-loop target-swap study, two learned targets attain higher observed success than the baseline on the LIBERO-10 object-position shift in single policy runs, but neither shows a clear gain in aggregate out-of-distribution success. These results support transition--action supervision as a source of action-relevant structure in the evaluated checkpoints, while showing that absolute probe performance is sensitive to task coverage and does not guarantee policy success.
\end{abstract}


\section{Introduction}

Robot policies and world-model-based planners increasingly reuse pretrained visual and video encoders~\citep{vjepa2, dinowm, openvla}. The implicit assumption is that a world model trained for next-frame prediction captures \emph{action-relevant} dynamical structure in its latent space, which a downstream policy can read out. This assumption matters because the choice of frozen-feature extractor is a load-bearing one for the rest of the pipeline: a representation that does not encode rotation, gripper transitions, or contact dynamics cannot be repaired by a stronger policy head. Two empirical observations complicate the assumption. The standard world-model evaluation metrics---PSNR, FVD, LPIPS---measure pixel-level prediction quality, and recent work~\citep{tian2023vp2, semantic_wm_2026} shows these correlate poorly with control utility, mirroring the longer-standing observation that perception-strong video models often fail at the underlying causal and dynamical structure of the same scene~\citep{clevrer}. At the same time, the field already spans a broad range of pretrained encoders---from reconstruction VAEs and masked image models to language-aligned vision encoders and latent prediction models---yet it remains unclear which pretraining signals actually induce action-relevant structure in the resulting latent space.

This paper asks when a video model's latent space carries action-relevant structure and how inverse-dynamics (ID) supervision changes it. We train or obtain ten backbone entries spanning seven objective families, optionally apply ID supervision following the principle of~\citet{wam}, and probe their features for action recovery on a fixed computational subset of the task-OOD split of LIBERO~\citep{libero}. A common probe and pixel decoder make checkpoint-level action readouts and pixel fidelity comparable; differences in training data and architecture still limit causal attribution across families. We evaluate both clean and visually perturbed inputs, and use all $26$ held-out tasks for the matched-control contrasts.

The results follow four questions. \emph{Which features carry action information?} On the fixed LIBERO subset, video-pretrained encoders recover actions more accurately than the highest-PSNR reconstruction models; rotation separates families most strongly, and a V-JEPA per-layer probe shows a mid-trunk plateau followed by a fall through the late blocks. \emph{What does ID add?} Matched action-shuffled and forward-dynamics controls over all held-out tasks show that correct transition--action pairing is needed for the large V-JEPA and VideoMAE relative gains, which also survive visual corruption. \emph{Where does the ranking hold?} It largely persists on MetaWorld, whereas image-only encoders close the gap in CALVIN's fixed environments. \emph{Does the probe predict control?} In a mostly single-seed $\pi_{0.5}$ target swap, learned targets attain higher observed LIBERO-10 position-shift success but show no clear aggregate LIBERO-PRO gain, so probe and policy results must be interpreted separately. Code is available at \url{https://github.com/k1seul/corl2026-action-relevant-latents}.

\section{The Action-Relevance Gap in Video World Model Latents}
\label{sec:related}
 
Prior work leaves three gaps that motivate our study: world models are often evaluated by pixel fidelity rather than action recoverability, action relevance is often measured only indirectly through downstream policy success, and many comparisons characterize a single architecture rather than pretraining objectives head-to-head.
 
\paragraph{World model evaluation and encoders for policy.} World model quality has historically been measured by pixel-prediction metrics (PSNR, FVD, LPIPS)~\citep{dreamerv3, iris}, with recent benchmarks~\citep{rbench, worldmodelbench, ewmbench, worldarena, worldinworld} extending evaluation to physics adherence and embodied utility. \citet{tian2023vp2} showed that perceptual metrics rank video predictors differently from control success; we extend this dissociation to ten backbone entries spanning seven objective families under a frozen-feature action probe. \citet{joseph2026physics} localize physical-variable representations inside V-JEPA-class encoders and report mid-depth peaks for temporally grounded signals; we observe an analogous mid-trunk action-probe profile, while the rotation separation is specific to our study. DINO-WM~\citep{dinowm} and V-JEPA $2$-AC~\citep{vjepa2} use frozen pretrained encoders as world-model substrates. \citet{semantic_wm_2026} compare six encoders as latents for action-conditioned diffusion on Bridge V2 by rolling out a frozen policy, whereas we probe the representation directly under a fixed protocol. Concurrent V-JEPA Policy~\citep{vjepa_policy_2026} compares predictive, discriminative, reconstructive, and video-understanding visual foundations within a shared downstream world-action model and reports predictive V-JEPA latents to be strongest, particularly under distribution shift. Our study instead measures action recoverability across a checkpoint matrix and isolates transition--action correspondence with matched shuffled-action and forward-dynamics controls. We also reuse the exact Cosmos-$1$ tokenizer of~\citet{semantic_wm_2026} and add pixel-diffusion and shortcut-forcing-dynamics architectures.
 
\paragraph{Inverse-dynamics supervision and representation robustness.} Predicting actions from consecutive frame features is a long-standing self-supervised signal~\citep{agrawal2016poke, pathak_icm, shelhamer2017, dynamo}. The closest prior to our recipe is WAM~\citep{wam}, which augments DreamerV2 with an inverse-dynamics head on CALVIN---a single architecture. We follow the same inverse-dynamics principle while using $L_2$ as our cross-architecture default, whereas WAM uses $L_1$. Under our protocol the objective produces a $+0.44$ $R^2$ lift on V-JEPA and $0.00$ on Dreamer $4$, a backbone dependence not previously documented. Separately, VLA policies are known to collapse under visual distribution shift~\citep{libero_pro}; on our fixed held-out subset, we provide the action-probe analog at the representation level, showing that frozen pretrained encoders are perturbation-sensitive and that inverse-dynamics supervision restores robustness in the same regimes where it lifts clean $R^2$.
 
\paragraph{Action-relevant representations and physical understanding.}
A line of work argues that strong perceptual representations need not encode the causal structure required for physical reasoning~\citep{clevrer, joseph2026physics, riochet2020intphysframeworkbenchmarkvisual}, probing whether predictive objectives align latent spaces with abstract physical variables like velocity or motion direction~\citep{joseph2026physics, tian2023vp2}. These probe physical variables in isolation rather than the action that generated a transition. Action-centric pretraining such as CAIP~\citep{caip} instead aligns image features with actions contrastively; we evaluate it as a comparator in Section~\ref{sec:matched}. We close that gap by asking whether pretrained video representations carry enough information to recover the underlying action, making inverse-dynamics recoverability a task-grounded definition of action-relevant structure. Across the evaluated checkpoints, temporal video pretraining is associated with stronger action readout than reconstruction-aligned pretraining; the unmatched comparison does not isolate the causal effect of the objective.

\section{Experimental Setup}
\label{sec:setup}

\paragraph{Benchmark.} LIBERO~\citep{libero} has $130$ tasks and a $7$-DoF action space (translation, rotation, gripper). We use its task-OOD split: $104$ training tasks and $26$ tasks held out entirely from world-model training. Unless stated otherwise, probes are trained on the first $400$ episodes of the training split, covering $10$ of the $104$ training tasks, and evaluated on the first $200$ episodes of the held-out split, covering $4$ of the $26$ held-out tasks ($2$ spatial and $2$ object tasks). Thus the absolute LIBERO probe values measure transfer to this fixed four-task subset rather than aggregate performance over all $26$ held-out tasks. The matched-control contrasts in Section~\ref{sec:matched} and Appendix~\ref{app:matched} use all $26$ held-out tasks. Cross-benchmark results on CALVIN~\citep{calvin} and MetaWorld~\citep{metaworld} are reported in Appendix~\ref{app:cross_benchmark}.

\paragraph{Encoders.} We compare ten backbone entries spanning seven objective families. \emph{DIFF} is a pixel-diffusion world model trained from scratch on LIBERO at $34$--$91$M parameters, used as the LIBERO-native baseline and for within-architecture ablations of $\lambda$, head form, and capacity. \emph{Dreamer 4}~\citep{dreamerv4} is an unofficial reproduction of the shortcut-forcing dynamics architecture; because the original implementation is not public, we build on the unofficial PyTorch implementation,\footnote{\url{https://github.com/nicklashansen/dreamer4}} training $64$M- and $276$M-parameter models from scratch on LIBERO and a $64$M model from that implementation's DMControl-pretrained checkpoint. We probe both agent-token and spatial-token feature streams and treat these results as exploratory. \emph{SDXL VAE}~\citep{sdxlvae} and the \emph{Cosmos-1 image tokenizer}~\citep{cosmos1} are reconstruction-aligned autoencoders; the Cosmos-1 checkpoint we use is the exact reconstruction baseline used in concurrent work~\citep{semantic_wm_2026}. \emph{LAPA}~\citep{lapa} is a discrete latent-action quantizer pretrained on Open X-Embodiment. \emph{Web-DINO}~\citep{webssl} is a ViT-L image encoder trained with masked contrastive prediction; \emph{SigLIP 2}~\citep{siglip2} is a ViT-L image-language contrastive encoder. \emph{VideoMAE}~\citep{videomae} V1 ViT-L is a video-pretrained encoder trained with pixel-space masked autoencoding. \emph{V-JEPA 2}~\citep{vjepa2} ViT-L and the distilled V-JEPA 2.1 ViT-B are video-pretrained encoders trained with feature-level masked latent prediction. For each pretrained encoder we report a frozen baseline and an ID-fine-tuned variant; for DIFF and Dreamer 4 we report from-scratch training with and without the ID loss. See Appendix~\ref{app:backbone_catalog} for checkpoint details.

\paragraph{Inverse-dynamics recipe.} Let $o_t$ denote the observation (RGB frames from the manipulator's camera rig at timestep $t$) and $a_t$ the corresponding $7$-DoF action ($3$ translation, $3$ rotation, $1$ gripper). For each encoder $f_\theta$, an inverse-dynamics head $g_\phi(f_\theta(o_t), f_\theta(o_{t+1})) \to \hat{a}_t$ is added during world-model training (DIFF, Dreamer 4) or fine-tuning (pretrained encoders), with loss $\mathcal{L} = \mathcal{L}_{\text{primary}} + \lambda \cdot \mathrm{MSE}(\hat{a}_t, a_t)$; for every pretrained-encoder row we fine-tune with the inverse-dynamics loss alone ($\mathcal{L}_{\text{primary}}{=}0$).
We chose this loss over three alternatives---forward dynamics, temporal contrastive, and single-frame action prediction---based on a head-form ablation on DIFF (Appendix~\ref{app:aux_ablations}); inverse dynamics produces the largest lift among the four under matched compute. We use the $L_2$ form as our cross-architecture default; WAM~\citep{wam} uses $L_1$, which performs within $0.05$ $R^2$ of $L_2$ on DIFF (Appendix~\ref{app:aux_ablations}). We use $\lambda{=}1.0$ for pretrained-encoder rows and $\lambda{=}0.05$ for DIFF; $\lambda$-sweeps over three orders of magnitude on DIFF and two on image-only SSL backbones show the choice does not change the ranking (Appendix~\ref{app:aux_ablations}). $g_\phi$ is discarded at probe time.

\paragraph{Probe.} We train an MLP $[D \to 256 \to 128 \to 7]$ for $3000$ steps (AdamW, lr $3 \times 10^{-4}$, batch $512$) on mean-pooled features of clean training frames. We report aggregate $R^2$ across the $7$ action dimensions and per-dimension $R^2$ where mechanism matters, averaged over $3$ probe seeds. Robustness uses the same probe with test-time perturbations on the trunk; see Section~\ref{sec:robustness}.

\section{Experimental Results}
\label{sec:results}

\begin{figure}[t]
    \centering
    \includegraphics[width=\linewidth]{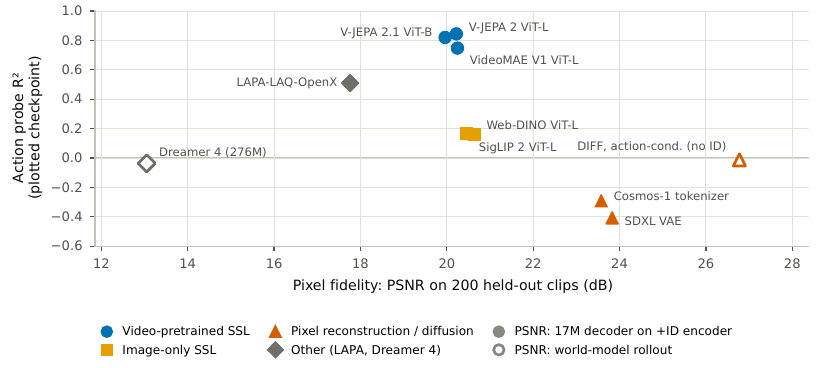}
    \caption{\textbf{Pixel fidelity and action recoverability across the evaluated checkpoints.} Each point is one checkpoint; the y-axis is action-probe $R^2$ on the fixed four-task LIBERO held-out subset. All points are $+$ID checkpoints except the action-conditioned DIFF, which has no ID loss. PSNR is measured on the same $200$ held-out clips: for encoder-only backbones, a $17$M-parameter pixel decoder is trained on top of the fixed $+$ID encoder (filled markers); for pixel-producing world models, it is rollout PSNR (hollow markers). The highest-PSNR models (the reconstruction autoencoders and action-conditioned DIFF) have $R^2\leq 0$, while the video-pretrained encoders reach the highest $R^2$ at moderate PSNR. The comparison is descriptive and does not establish a correlation or a Pareto frontier.}
    \label{fig:pareto}
\end{figure}

\subsection{Does Pixel Fidelity Predict Action-Relevant Structure?}
\label{sec:orthogonal}

The standard world-model selection criterion is pixel-prediction quality (PSNR, FVD, LPIPS). We ask whether this criterion is a useful proxy for action-relevant representation quality. To place encoder-only backbones on the same axes as pixel-producing backbones, we train a lightweight $17$M-parameter pixel decoder ($4$-stage ConvTranspose, MSE $+$ LPIPS loss, $10$k steps) on top of each fixed $+$ID encoder; the decoder is used only to produce a PSNR measurement comparable across families. Figure~\ref{fig:pareto} plots the representative checkpoint from each backbone family on the PSNR-vs-action-$R^2$ plane. The action-conditioned DIFF point has no ID loss and uses rollout PSNR.

The central observation of Figure~\ref{fig:pareto} is that, across the checkpoints we test, pixel prediction quality is not a reliable proxy for action recoverability on the fixed subset. Five checkpoints between $19.96$ and $20.63$ dB span $+$ID action $R^2$ from $0.16$ to $0.84$, while the checkpoints with $R^2\leq 0$ vary by almost $14$ dB in PSNR. The action-conditioned DIFF and reconstruction autoencoders achieve the highest PSNR yet have non-positive action $R^2$, whereas the V-JEPA checkpoints retain strong action readouts despite moderate pixel fidelity. Because data, architecture, and adaptation differ across points, this is a checkpoint-level diagnostic rather than a causal estimate of objective effects.

Separately, the frozen values in Table~\ref{tab:multiplier} do not cleanly separate the strongest families on the same subset: video-pretrained SSL, pixel diffusion, and latent-action quantization cluster within $R^2 \in [0.40, 0.46]$ despite different pretraining objectives. Their separation after inverse-dynamics supervision indicates different adaptation behavior, but the unmatched cross-family comparison cannot isolate pretraining objective from data, architecture, tokenization, or scale. Section~\ref{sec:matched} therefore provides the controlled evidence over all $26$ held-out tasks.

\subsection{Does the Inverse-Dynamics Loss Lift Every Backbone Equally?}
\label{sec:multiplier}

\begin{table}[t]
    \centering
    \footnotesize
    \setlength{\tabcolsep}{3pt}
    \resizebox{\linewidth}{!}{%
    \begin{tabular}{l l c l c c c c}
    \toprule
    Backbone & Family & Lang. & Pretrain data & Params & Frozen & $+$ ID & $\Delta$ \\
    \midrule
    \midrule
    V-JEPA 2 ViT-L          & \multirow{2}{*}{Video $+$ JEPA pred.} & \multirow{2}{*}{--} & \multirow{2}{*}{Internet video}    & $304$M & $0.40$  & $\textbf{0.84}$           & $\textbf{+0.44}$ \\
    V-JEPA 2.1 ViT-B        &                                       &                     &                                    & $87$M  & $0.44$  & $0.82$                    & $+0.38$ \\
    \midrule
    VideoMAE V1 ViT-L       & Video $+$ Pixel MAE                   & --                  & Internet video                     & $304$M & $0.46$  & $0.75$                    & $+0.29$ \\
    \midrule
    Web-DINO ViT-L          & \multirow{2}{*}{Image SSL}            & --                  & Web images                         & $304$M & $-0.01$ & $0.16$                    & $+0.17$ \\
    SigLIP 2 ViT-L          &                                       & $\checkmark$        & Web img-text                       & $316$M & $0.05$  & $0.17$                    & $+0.12$ \\
    \midrule
    LAPA-LAQ-OpenX          & Latent action quant.                      & --                  & Open X-Embod.                      & $344$M & $0.41$  & $0.51$                    & $+0.10$ \\
    \midrule
    DIFF                    & Pixel diffusion                       & --                  & LIBERO 104-task                    & $91$M  & $0.43$  & $0.57$                    & $+0.15$ \\
    \midrule
    SDXL VAE                & \multirow{2}{*}{Pixel reconstruction} & \multirow{2}{*}{--} & \multirow{2}{*}{Web images}        & $34$M  & $-0.55$ & $-0.41$                   & $+0.14$ \\
    Cosmos-$1$ tokenizer    &                                       &                     &                                    & $34$M  & $-0.36$ & $-0.29$                   & $+0.07$ \\
    \midrule
    Dreamer 4               & Shortcut dyn.                         & --                  & LIBERO 104-task                    & $64$--$276$M & $-0.04$ & $-0.04$                   & $0.00$ \\
    \bottomrule
    \end{tabular}}
    \caption{\textbf{Inverse-dynamics auxiliary lift, ordered by pretraining objective family.} Absolute $R^2$ values use the fixed four-task LIBERO held-out subset. All backbones share the inverse-dynamics objective and probe protocol of Section~\ref{sec:setup}; optimization settings differ by backbone and are reported in Appendix~\ref{app:training_settings}. Per-variant detail is in Appendix~\ref{app:full_multiplier_table}.}
    \label{tab:multiplier}
\end{table}

Table~\ref{tab:multiplier} shows that ID supervision has checkpoint-dependent effects: video-pretrained encoders improve substantially, whereas image-only and reconstruction checkpoints remain limited. Section~\ref{sec:matched} tests whether the lift requires correct transition--action pairing.

On the fixed subset, V-JEPA $2$ $+$ ID reaches $0.84$ R², VideoMAE $+$ ID $0.75$, and Web-DINO $+$ ID $0.16$. The video-pretrained models have a large advantage over image-only SSL; V-JEPA exceeds VideoMAE by $0.10$ here and $0.03$ in the separate matched-control protocol of Table~\ref{tab:matched}. Because their pretraining data and objectives are not jointly controlled, this difference is consistent with an additional benefit from feature-level prediction, rather than an isolated estimate of its effect.

Image-only SSL improves over the reconstruction checkpoints after ID tuning but remains far below the video-pretrained encoders, reaching only $0.16$--$0.17$ R². A $\lambda$ sweep over two orders of magnitude ($0.01$--$1.0$) leaves Web-DINO and SigLIP $2$ in a narrow $0.1$-wide R² band (Appendix~\ref{app:aux_ablations}), so the gap is not explained by the tested inverse-dynamics weight alone. These checkpoint results therefore do not support grouping V-JEPA with image-only ``semantic SSL'' methods under this protocol.

Model capacity alone also does not account for the ranking among the checkpoints we test. At $\sim$$90$M parameters, V-JEPA $2.1$ ViT-B $+$ ID ($0.82$) exceeds DIFF $+$ ID ($0.57$) by $0.25$ R². Among the $+$ID Dreamer $4$ runs, the $276$M model also remains at the $-0.04$ floor; there is no frozen $276$M baseline, and because this is an unofficial reproduction, we treat these results as exploratory. The evaluated capacity differences therefore do not explain the observed ranking.

The ordering largely persists on MetaWorld (Spearman $\rho{=}+0.88$ against LIBERO), but image-only SSL narrows the gap on CALVIN ($0.77$--$0.81$ R²). One plausible explanation is that CALVIN's four fixed tabletop environments allow static appearance to substitute for temporal context; Appendix~\ref{app:cross_benchmark} gives the cross-benchmark results and probe-budget control.

\subsection{Is the Lift Driven by Transition--Action Correspondence?}
\label{sec:matched}

\begin{table}[t]
    \centering
    \small
    \setlength{\tabcolsep}{4pt}
    \begin{tabular}{l c c c c c}
    \toprule
    Encoder & Frozen & True ID & Action-shuf.\ ID & FD & $\Delta_{26}$ (True $-$ shuf.) \\
    \midrule
    V-JEPA 2 ViT-L    & $0.455$ & $\mathbf{0.863}$ & $0.184$ & $0.326$  & $0.836$ [$0.752$, $0.952$] \\
    VideoMAE V1 ViT-L & $0.513$ & $0.829$          & $0.422$ & $0.454$  & $0.578$ [$0.518$, $0.658$] \\
    Web-DINO ViT-L    & $0.075$ & $0.156$          & $0.063$ & $-0.071$ & $0.097$ [$0.026$, $0.156$] \\
    \bottomrule
    \end{tabular}
    \caption{\textbf{Matched controls.} Action $R^2$ after $20$k updates from the same public checkpoint within each backbone. Absolute columns use the fixed $200$-episode subset and average two training seeds $\times$ three probe seeds (frozen: three probe seeds); frozen values were re-evaluated with the recovered submitted evaluator in the rebuttal environment and are within $0.10$ of Table~\ref{tab:multiplier}. $\Delta_{26}$ pools all $26$ held-out tasks and reports $95\%$ paired held-out-task cluster-bootstrap CIs ($10$k replicates), conditional on the two trained checkpoints and three probe seeds; it is not the arithmetic difference between the absolute columns.}
    \label{tab:matched}
\end{table}

To distinguish action correspondence from adaptation, we fine-tune identical V-JEPA $2$, VideoMAE, and Web-DINO checkpoints under true ID, within-task action-shuffled ID, or forward dynamics (FD), with the update budget matched within each backbone. The shuffled control breaks transition--action pairing; FD predicts $\mathrm{sg}(z_{t+1})$ from $(z_t,a_t)$. All training heads are discarded before fresh action probes are fitted (Appendix~\ref{app:matched}).

True ID outperforms action-shuffled ID for V-JEPA $2$ and VideoMAE ($\Delta_{26}=0.84$ and $0.58$), including under the worst visual corruption (gaps $0.66$ and $0.69$); Web-DINO's clean gap is only $0.10$ and vanishes under corruption. FD does not recover the gains. Correct correspondence thus matters for these video checkpoints, while ID is not a universal multiplier. VideoMAE approaches V-JEPA $2$ under this protocol ($0.829$ vs.\ $0.863$).

An action-centric comparator, the official CAIP ViT-L checkpoint~\citep{caip}, reaches frozen $R^2=0.019$ on the fixed subset, versus $0.449$ for V-JEPA $2$ and $-0.024$ for a constant predictor under the same separate encoding pass. Appendix~\ref{app:matched} reports both this comparison and the all-$26$-task coverage check. This checkpoint-level result does not characterize action-centric pretraining generally.

\begin{figure*}[t]
    \centering
    \includegraphics[width=\textwidth]{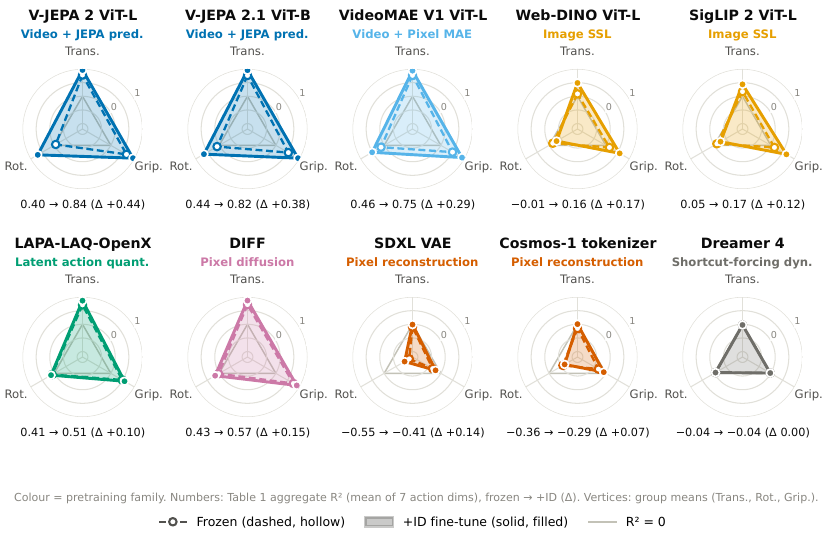}
    \caption{\textbf{Per-dimension action $R^2$ by pretraining backbone.}
    Vertices are translation, rotation, and gripper $R^2$ (group means of the per-dimension $R^2$); the gray triangle marks $R^2=0$.
    Solid filled triangles are $+$ID fine-tunes and dashed hollow triangles are frozen baselines.
    Numbers below each panel are the Table~\ref{tab:multiplier} aggregate (mean of the seven per-dimension $R^2$), frozen $\rightarrow$ $+$ID, and their difference $\Delta$.
    The DIFF panel uses the canonical $L_2$ variant of Table~\ref{tab:multiplier}. Mean of $3$ probe seeds on the fixed four-task held-out subset.}
    \label{fig:perdim}
\end{figure*}

\begin{figure}[t]
    \centering
    \includegraphics[width=\linewidth]{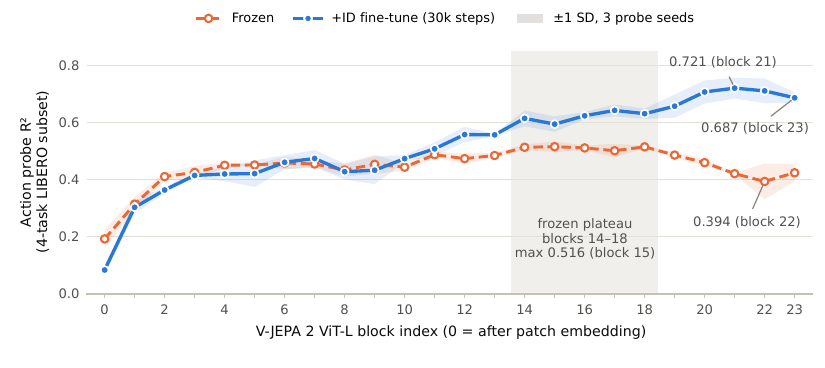}
    \caption{\textbf{V-JEPA $2$ ViT-L per-block action $R^2$.} Mean $\pm 1$ SD over $3$ probe seeds (one checkpoint per curve); features are block outputs taken before the encoder's final LayerNorm. The frozen trunk plateaus over blocks $14$--$18$ (maximum $0.516$ at block $15$) and falls to $0.394$ at block $22$; the ID fine-tune ($30$k steps) lifts the last four blocks by $+0.25$ to $+0.32$ $R^2$ and peaks at block $21$ ($0.721$). Full values in Table~\ref{tab:vjepa_perlayer}.}
    \label{fig:vjepa_perlayer}
\end{figure}

\subsection{Where Does the V-JEPA Advantage Come From?}
\label{sec:rotation}

Figure~\ref{fig:perdim} reveals that the dominant bottleneck is end-effector orientation, not translation. Translation and gripper state are recoverable even from relatively weak visual features, while rotation sharply separates predictive video models from appearance-driven encoders: in the radar plots the ``Rot.'' axis collapses outside the video-predictive family, and only V-JEPA sustains high performance across all three dimension groups simultaneously.

The appearance-driven families illustrate this asymmetry. Web-DINO and SigLIP retain moderate translation but produce negative rotation even after inverse-dynamics supervision, a pattern consistent with coarse appearance cues supporting translation more than viewpoint-consistent 3D orientation~\citep{joseph2026physics}. The reconstruction checkpoints (SDXL VAE, Cosmos-1) show the same rotation-specific collapse while preserving more translation, but this checkpoint comparison does not isolate the objective or the encoded geometry. Dreamer~4 instead floors near zero on all three groups.

Compared with the image-only SSL checkpoints, VideoMAE has higher observed rotational decoding. This is consistent with a benefit from temporal video pretraining, but the unmatched comparison does not isolate temporal exposure from data, architecture, or objective. V-JEPA expands the rotation axis further in the evaluated checkpoints while translation and gripper remain comparable, so its observed margin is concentrated on rotation.

The per-layer diagnostic in Figure~\ref{fig:vjepa_perlayer} shows a broad frozen plateau over blocks $14$--$18$ (maximum at block $15$), followed by a fall through the late blocks; ID fine-tuning lifts the final four blocks by $0.25$--$0.32$ and moves the maximum to block $21$. Along with the sparse-supervision sweep (Appendix~\ref{app:id_sparsity}), this pattern is consistent with ID making existing action information more accessible. It does not isolate the effect of V-JEPA's predictor during pretraining.

\subsection{Does the Inverse-Dynamics Lift Survive Distribution Shift?}
\label{sec:robustness}

\begin{table}[t]
    \centering
    \small
    \setlength{\tabcolsep}{4pt}
    \begin{tabular}{l r r r r r r}
    \toprule
    & \multicolumn{3}{c}{Frozen} & \multicolumn{3}{c}{$+$ ID} \\
    \cmidrule(lr){2-4}\cmidrule(lr){5-7}
    Backbone & clean & noise $0.15$ & worst & clean & noise $0.15$ & worst \\
    \midrule
    V-JEPA 2 ViT-L     & $0.40$  & $-1.18$ & $-4.66$ & $\mathbf{0.84}$ & $\mathbf{0.66}$ & $\mathbf{0.56}$ \\
    V-JEPA 2.1 ViT-B   & $0.44$  & $-5.36$ & $-5.36$ & $0.82$  & $0.01$  & $0.01$ \\
    VideoMAE V1 ViT-L  & $0.46$  & $-2.54$ & $-2.54$ & $0.75$  & $0.11$  & $0.11$ \\
    Web-DINO ViT-L     & $-0.01$ & $-4.19$ & $-4.19$ & $0.16$  & $0.08$  & $-0.05$ \\
    SigLIP 2 ViT-L     & $0.05$  & $-0.62$ & $-1.75$ & $0.17$  & $-0.09$ & $-0.40$ \\
    LAPA-LAQ-OpenX     & $0.41$  & $-1.14$ & $-1.14$ & $0.51$  & $-0.19$ & $-0.19$ \\
    DIFF               & $0.43$  & $0.44$  & $0.24$  & $0.57$  & $0.57$  & $0.46$ \\
    SDXL VAE           & $-0.55$ & $-1.96$ & $-1.96$ & $-0.41$ & $-2.20$ & $-2.20$ \\
    Cosmos-1 tokenizer & $-0.36$ & $-0.80$ & $-0.94$ & $-0.29$ & $-0.86$ & $-1.09$ \\
    Dreamer 4          & $-0.04$ & $-0.04$ & $-0.04$ & $-0.04$ & $-0.04$ & $-0.04$ \\
    \bottomrule
    \end{tabular}
    \caption{\textbf{Robustness to visual perturbation on the fixed four-task held-out subset.} The probe is trained on clean features and evaluated on perturbed inputs. ``Worst'' is the minimum over seven conditions: Gaussian noise (std $0.05$, $0.10$, $0.15$), color jitter ($0.10$, $0.20$), and Gaussian blur (kernel $5$, $11$). Frozen DIFF is in-distribution for LIBERO, and Dreamer 4 floors at $-0.04$ throughout. Per-condition results in Table~\ref{tab:perturb_full}.}
    \label{tab:perturb}
\end{table}

On the fixed subset, frozen pretrained encoders are highly perturbation-sensitive: frozen V-JEPA $2$ ViT-L falls from clean $0.40$ to $-4.66$ at its worst condition and frozen VideoMAE to $-2.54$, with similar collapses for Web-DINO and SigLIP. The corresponding ID-fine-tuned video checkpoints degrade more gracefully (V-JEPA $2$ ViT-L $+$ ID $0.84 \rightarrow 0.56$; VideoMAE $+$ ID remains positive). This establishes a robustness difference for the tested checkpoints but does not by itself identify the representation mechanism that causes it.

V-JEPA $+$ ID is more noise-robust than VideoMAE $+$ ID (noise $0.15$: $0.66$ vs.\ $0.11$), a pattern consistent with predictive latent objectives being less pixel-tied than masked pixel reconstruction, without isolating the objective from other checkpoint differences. SDXL VAE $+$ ID and Cosmos $+$ ID instead become \emph{more} negative as perturbation strengthens, while Dreamer~4 stays flat at its information floor. The matched controls (Section~\ref{sec:matched}) show that the True$-$shuffled robustness contrast also depends on correct transition--action pairing.

\subsection{Do Action-Relevant Targets Help Closed-Loop Control?}
\label{sec:closedloop}

\begin{table}[t]
    \centering
    \footnotesize
    \setlength{\tabcolsep}{3pt}
    \begin{tabular}{l c c c c c c c c c c c}
    \toprule
    & \multicolumn{5}{c}{LIBERO (in distribution)} & \multicolumn{5}{c}{LIBERO-PRO, L10 shifts} & PRO \\
    \cmidrule(lr){2-6}\cmidrule(lr){7-11}\cmidrule(lr){12-12}
    Auxiliary target & Spat. & Obj. & Goal & L10 & Avg. & Lang. & Pos. & Obj. & Task & Env. & All \\
    \midrule
    None ($\pi_{0.5}$ baseline)        & $96$ & $99$ & $97$ & $80$ & $93.0$ & $65$ & $8$ & $79$ & $15$ & $47$ & $68.7$ \\
    \midrule
    \multicolumn{12}{l}{\emph{Clip coupling} (normalized $L_2$, weight $0.1$)} \\
    DIFF $+$ ID$^{\dagger}$            & $96$ & $98$ & $93$ & $88$ & $\mathbf{93.8}$ & $63$ & $\mathbf{24}$ & $85$ & $16$ & $55$ & $\mathbf{69.0}$ \\
    \quad batch-shuffled target        & $96$ & $99$ & $83$ & $78$ & $89.0$ & $65$ & $15$ & $81$ & $15$ & $60$ & $65.8$ \\
    \quad random-init encoder          & $95$ & $98$ & $94$ & $87$ & $93.5$ & $72$ & $14$ & $82$ & $25$ & $48$ & $67.6$ \\
    \midrule
    \multicolumn{12}{l}{\emph{Token coupling} (cosine)} \\
    V-JEPA 2 $+$ ID (weight $0.01$)    & $95$ & $98$ & $90$ & $78$ & $90.2$ & $54$ & $23$ & $79$ & $19$ & $55$ & $67.5$ \\
    Web-DINO (weight $0.01$)           & $99$ & $98$ & $91$ & $86$ & $93.5$ & $56$ & $14$ & $83$ & $16$ & $54$ & $65.0$ \\
    \bottomrule
    \end{tabular}
    \caption{\textbf{Closed-loop target swap} (success rate, \%). Policy initialization, data, and update budget are fixed; target couplings differ between the two groups. LIBERO: $10$ tasks $\times$ $10$ trials per suite. LIBERO-PRO: the five shifts on the long-horizon LIBERO-10 suite, and ``All'' averages the five shifts over all four suites. All displayed rows report policy seed $0$. $^{\dagger}$A second DIFF $+$ ID seed, evaluated only on LIBERO-PRO, reaches Pos. $18$ and PRO All $65.1$. Bold is restricted to the best observed value in the three headline columns: LIBERO Avg., L10 Pos., and PRO All; these are descriptive single-run comparisons without uncertainty estimates. Full per-suite results and the V-JEPA weight-$0.05$ sensitivity run are in Appendix~\ref{app:closedloop}.}
    \label{tab:closedloop}
\end{table}

To examine the link between probe readouts and control, we fine-tune $\pi_{0.5}$~\citep{pi05} on LIBERO~\citep{libero} with an auxiliary target from a frozen encoder. Policy initialization, data, and update budget are fixed. The \emph{clip} coupling uses the DIFF $+$ ID configuration corresponding to Table~\ref{tab:multiplier} (probe $R^2=0.57$) and compares it with batch-shuffled and random-initialized controls; this correspondence is established by the saved checkpoint configuration and run path rather than a byte-level weight hash. The \emph{token} coupling compares a LIBERO ID-fine-tuned V-JEPA $2$ target with the public frozen Web-DINO checkpoint. These couplings also differ in loss and weight, so their results are not a controlled comparison of target encoders across groups (Appendix~\ref{app:closedloop}).

On the LIBERO-10 position shift, the DIFF $+$ ID and V-JEPA $2$ target runs reach $24\%$ and $23\%$, versus $8\%$ for the baseline and $14$--$15\%$ for the controls; the second DIFF $+$ ID policy seed reaches $18\%$ (Appendix~\ref{app:closedloop}). The batch-shuffled target leads on the environment shift ($60\%$), and none of the targets clearly exceeds the aggregate LIBERO-PRO mean of the baseline ($65.0$--$69.0\%$ vs.\ $68.7\%$). The higher observed position-shift success is therefore specific to this suite and setup. Raising the V-JEPA $2$ target weight to $0.05$ removes it ($8\%$; Appendix~\ref{app:closedloop}). Probe $R^2$ also does not monotonically rank the policies, so action readout and closed-loop success should be evaluated separately.

\section{Limitations}
\label{sec:limitations}

All three benchmarks are simulated, and transfer to real robots is untested. The primary LIBERO cross-encoder comparisons use a fixed four-task subset of the $26$ held-out tasks because full-coverage $+$ID evaluation was not available. The all-$26$-task frozen coverage check in Table~\ref{tab:caip_coverage} substantially lowers absolute $R^2$ (V-JEPA $2$: $0.449\rightarrow-0.006$; VideoMAE: $0.472\rightarrow-0.186$), and frozen V-JEPA $2$ is not distinguishable from the constant predictor under its task-cluster CI. The absolute values should therefore not be interpreted as benchmark-wide action-recovery estimates. Our broader relative conclusion relies on the matched-control $\Delta_{26}$ comparison, which uses all $26$ held-out tasks. Public and reproduced checkpoints also differ in data, architecture, and scale, so the cross-family and CAIP comparisons describe these checkpoints rather than isolated pretraining effects; the PSNR comparison is descriptive. The closed-loop study uses few targets, two couplings, and mostly one policy seed: its LIBERO-10 position-shift gain is not an aggregate LIBERO-PRO gain. Dreamer $4$ is an unofficial reproduction, and the linear action-subspace (Ridge) analysis is exploratory. Real-robot, full-coverage $+$ID, and multi-seed policy studies remain future work.

\section{Conclusion}
\label{sec:conclusion}

On the fixed four-task LIBERO subset, pixel fidelity was an unreliable proxy for action recoverability, and video-pretrained encoders offered stronger action readouts, especially for end-effector rotation. Across all $26$ held-out tasks, matched controls show that the relative V-JEPA $2$ and VideoMAE gains from ID depend on correct transition--action pairing. Absolute frozen recoverability is task-coverage-sensitive, so these claims should not be read as benchmark-wide action-recovery estimates. A limited closed-loop target swap shows higher observed LIBERO-10 position-shift success for two learned targets but not a clear aggregate LIBERO-PRO gain. These findings favor testing action-relevant representations directly while keeping absolute probe quality, relative controlled effects, and policy success distinct.

\acknowledgments{This research was supported by a grant of the Korea Health Technology R\&D Project through the Korea Health Industry Development Institute (KHIDI), funded by the Ministry of Health \& Welfare, Republic of Korea (grant number: RS-2025-02307233). This work was supported by Institute of Information \& communications Technology Planning \& Evaluation (IITP) under the Leading Generative AI Human Resources Development (IITP-2026-RS-2024-00397085) grant funded by the Korea government (MSIT). This work was supported by the National Research Foundation of Korea (NRF) grant funded by the Korea government (MSIT) (No. RS-2023-00222663).}

\bibliography{main}

\newpage

\appendix

\section{Inverse-Dynamics Loss Ablations on DIFF}
\label{app:aux_ablations}

All cross-architecture rows in the main text use the same inverse-dynamics recipe: a single inverse-dynamics head $g_\phi(f_\theta(o_t), f_\theta(o_{t+1})) \to \hat{a}_t$ trained with $L_2$ MSE, at $\lambda{=}1.0$ for pretrained-encoder rows and $\lambda{=}0.05$ for DIFF. This appendix reports the ablations behind these choices, conducted on the canonical DIFF backbone ($91$M parameters, $2$-camera input, no action conditioning, task-OOD training split) where the hyperparameter sweeps were cheap enough to run exhaustively.

\paragraph{$\lambda$ sweep.} We swept the inverse-dynamics weight $\lambda$ across three orders of magnitude (Table~\ref{tab:lambda_sweep}). Action $R^2$ follows an inverted-U with a plateau between $\lambda{=}0.01$ ($R^2$ $0.598$) and $\lambda{=}0.05$ ($R^2$ $0.572$): the two settings differ by only $0.026$ in $R^2$ and are respectively $+0.17$ and $+0.15$ above the no-ID baseline ($0.43$). Beyond $\lambda{=}0.1$, both action readout and generation quality degrade: action $R^2$ falls below $0.3$ at $\lambda\geq0.3$, while rollout PSNR drops by $5$--$9$ dB only at $\lambda\geq1.0$. Recon MSE and LPIPS move smoothly across the plateau (within $0.0007$ and $0.06$ respectively), so the ID loss's effect on generation quality is small in the useful range. Appendix~\ref{app:cosine} pairs this $\lambda$ sweep with a temporal-cosine diagnostic.

\begin{table}[h]
    \centering
    \small
    \begin{tabular}{c c c c c}
    \toprule
    $\lambda$ & Action $R^2$ & Rollout PSNR & Recon MSE & LPIPS \\
    \midrule
    ---       & $0.426$            & $19.18$            & $0.0121$ & $0.287$ \\
    $0.01$    & $\textbf{0.598}$   & $19.07$ ($-0.11$)  & $0.0124$ & $0.297$ \\
    $0.025$   & $0.571$            & $19.03$ ($-0.15$)  & $0.0125$ & $0.320$ \\
    $0.05$    & $0.572$            & $18.91$ ($-0.27$)  & $0.0128$ & $0.344$ \\
    $0.1$     & $0.436$            & $18.54$ ($-0.65$)  & $0.0140$ & $0.386$ \\
    $0.15$    & $0.306$            & $18.12$ ($-1.06$)  & $0.0154$ & $0.424$ \\
    $0.3$     & $0.282$            & $17.40$ ($-1.79$)  & $0.0182$ & $0.485$ \\
    $1.0$     & $0.264$            & $13.75$ ($-5.43$)  & $0.0421$ & $0.622$ \\
    $10$      & $0.145$            & $10.62$ ($-8.57$)  & $0.0867$ & $0.718$ \\
    \bottomrule
    \end{tabular}
    \caption{\textbf{Auxiliary weight $\lambda$ sweep on canonical DIFF.} Action $R^2$ is inverted-U with a plateau at $\lambda \in [0.01, 0.05]$. We use $\lambda{=}0.05$ for the DIFF rows of the main text; pretrained-encoder rows use $\lambda{=}1.0$. The $\lambda{=}0.01$ DIFF variant would yield $\sim$$0.03$ higher absolute $R^2$ but would not change any qualitative ranking.}
    \label{tab:lambda_sweep}
\end{table}

\paragraph{Auxiliary loss form.} We hold $\lambda{=}0.05$ fixed and vary the auxiliary loss: $L_2$ vs.\ $L_1$ inverse-dynamics regression, forward dynamics, temporal contrastive, and single-frame action prediction (Table~\ref{tab:aux_head}). Both inverse-dynamics regressions sit at the top; $L_1$ regression (matching WAM's original specification) outperforms our $L_2$ default by $+0.05$ on action $R^2$ ($0.629$ vs.\ $0.578$) at essentially identical rollout PSNR ($18.84$ vs.\ $18.85$ dB). Single-frame action prediction ($g_\phi(f_t) \to a_t$, no $f_{t+1}$) drops to $R^2$ $0.523$, confirming that consecutive-frame structure is what the loss depends on. Forward dynamics ($g_\phi(f_t, a_t) \to f_{t+1}$) falls to $0.408$, suggesting the direction of inversion matters: predicting actions from feature pairs shapes the trunk more effectively than predicting features from action-conditioned pairs. A temporal-contrastive objective collapses to $0.073$.

\begin{table}[h]
    \centering
    \small
    \begin{tabular}{l c c}
    \toprule
    Loss form & Action $R^2$ & Rollout PSNR \\
    \midrule
    $L_1$ inverse dynamics                       & $\textbf{0.629}$ & $18.84$ \\
    $L_2$ MSE inverse dynamics (default)         & $0.578$          & $18.85$ \\
    Single-frame action prediction               & $0.523$          & $18.85$ \\
    Forward dynamics ($f_t, a_t \to f_{t+1}$)    & $0.408$          & $19.11$ \\
    Temporal contrastive                         & $0.073$          & $18.43$ \\
    \bottomrule
    \end{tabular}
    \caption{\textbf{Auxiliary loss-form ablation at $\lambda{=}0.05$.} The head architecture is fixed; only the loss form varies. $L_1$ inverse-dynamics regression is the best objective by $+0.05$ over the $L_2$ default. We retain $L_2$ as the canonical setting in the main text because it is the more common choice and to avoid conflating ``ID supervision helps'' with ``a particular loss form helps''; the $L_1$ variant would shift the DIFF $+$ ID row in Table~\ref{tab:multiplier} from $0.57$ to $0.63$.}
    \label{tab:aux_head}
\end{table}

\paragraph{Prediction horizon $k$.} Extending the ID loss from $1$-step ($g_\phi(f_t, f_{t+1}) \to a_t$) to $k$-step ($g_\phi(f_t, f_{t+k}) \to (a_t, \dots, a_{t+k-1})$) traces an inverted-U with peak at $k{=}4$: action $R^2$ moves from $0.578$ ($k{=}1$) to $0.587$ ($k{=}2$) to $\mathbf{0.608}$ ($k{=}4$) to $0.546$ ($k{=}8$). The peak at $k{=}4$ aligns with the typical action chunk length used by downstream chunk-style action heads~\citep{rt2, openvla, pi0}. We retain $k{=}1$ as the canonical setting for direct comparability with WAM.

\paragraph{Discussion.} The ablations support two narrower claims about the ID loss. First, the recipe is forgiving in a band around the canonical settings ($\lambda \in [0.01, 0.05]$, $L_2$ or $L_1$, $k \in [1, 4]$); none of these knobs is fragile within the tested range. Second, the cross-backbone spread in Section~\ref{sec:multiplier} ($+0.44$ on V-JEPA vs.\ $0.00$ on Dreamer 4) is larger than the variations observed among the tested DIFF settings ($+0.05$ at most), suggesting that it is not explained by the canonical DIFF recipe choice alone.

\paragraph{$\lambda$ sweep on image-only SSL backbones.} A natural concern is whether the V-JEPA-specific multiplier in Section~\ref{sec:multiplier} (Web-DINO $+$ ID $0.16$, SigLIP 2 $+$ ID $0.17$ at $\lambda{=}1.0$) is an artifact of the inverse-dynamics weight setting. We rule this out by sweeping $\lambda \in \{0.01, 0.025, 0.1, 0.3, 1.0\}$ on both Web-DINO ViT-L and SigLIP 2 ViT-L, with all other settings held fixed (Table~\ref{tab:image_ssl_lambda}).

\begin{table}[h]
    \centering
    \small
    \begin{tabular}{c c c}
    \toprule
    $\lambda$ & Web-DINO test $R^2$ & SigLIP 2 test $R^2$ \\
    \midrule
    $0.01$  & $0.134$ & $0.143$ \\
    $0.025$ & $0.122$ & $0.106$ \\
    $0.1$   & $\mathbf{0.168}$ & $0.139$ \\
    $0.3$   & $0.076$ & $0.077$ \\
    $1.0$ (main text) & $0.159$ & $\mathbf{0.166}$ \\
    \bottomrule
    \end{tabular}
    \caption{\textbf{ID $\lambda$ sweep on image-only SSL backbones.} Both Web-DINO and SigLIP 2 stay in a $0.10$-wide band ($R^2 \in [0.07, 0.17]$) across all $\lambda$ values tested. The best per backbone is only $\sim$$0.01$ above the $\lambda{=}1.0$ main-text setting, and no tested $\lambda$ closes the gap to V-JEPA $+$ ID at $0.84$. Both show a dip at $\lambda{=}0.3$, but without repeated runs we cannot separate optimization instability from run variance.}
    \label{tab:image_ssl_lambda}
\end{table}

\section{DIFF-Specific Findings: Action Conditioning and Capacity}
\label{app:diff_findings}

This appendix reports two diagnostic findings on our DIFF pixel-diffusion backbone that informed how we set up the cross-architecture rows in Section~\ref{sec:multiplier}, but are not central to the cross-architecture claim itself.

\paragraph{The action-conditioning paradox.} A natural baseline for a robot world model is to condition the trunk on the action sequence at training time, since the model is then explicitly modeling action-conditioned futures. We find that this hurts action recovery from the resulting frozen features. The first two rows of Table~\ref{tab:paradox} form the controlled comparison: they differ only in whether action conditioning is provided to the trunk. The remaining rows are diagnostic variants that also change training length, capacity, camera setup, or split.

\begin{table}[h]
    \centering
    \small
    \begin{tabular}{l c c c c c}
    \toprule
    Variant & cam & action input & steps & size & Action $R^2$ \\
    \midrule
    DIFF (no action)        & $1$ & ---         & $100$k & medium & $+0.26$ \\
    DIFF + action           & $1$ & \checkmark  & $100$k & medium & $-0.32$ \\
    DIFF + action, longer   & $1$ & \checkmark  & $200$k & medium & $+0.06$ \\
    DIFF + action, XL       & $1$ & \checkmark  & $200$k & XL     & $-0.01$ \\
    DIFF (no action), XL    & $1$ & ---         & $100$k & XL     & $+0.31$ \\
    DIFF (no action), $2$cam & $2$ & ---        & $100$k & XL     & $+0.36$ \\
    DIFF (no action), $2$cam, task-OOD & $2$ & --- & $100$k & XL & $+0.43$ \\
    \bottomrule
    \end{tabular}
    \caption{\textbf{Action conditioning hurts action recovery in the matched DIFF pair.} At medium capacity, one camera, and $100$k steps, adding action input drops $R^2$ by $0.58$ ($0.26 \to -0.32$). The remaining rows diagnose longer training, larger capacity, a second camera, and the task-OOD split; because these changes are sequential rather than fully factorial, they should not be read as isolated effects.}
    \label{tab:paradox}
\end{table}

One possible explanation is that, when the trunk receives the action sequence as input, it can route action-dependent prediction through the conditioning pathway and leave less action-relevant information in the visual features. The matched pair is consistent with this account, but the probe result alone does not establish that routing mechanism.

This is a within-DIFF observation; we do not claim it generalizes to other architectures (V-JEPA, Web-DINO, etc., do not see action data at pretraining and so cannot exhibit this paradox in the same form). It does, however, motivate our cross-architecture choice to never feed actions to the trunk for the no-ID baselines: doing so would systematically suppress the very signal the probe measures.

\paragraph{Capacity and camera setup.} At one camera without action input, moving from the medium model ($0.26$) to XL ($0.31$) yields a small observed increase. Adding a wrist camera to the XL configuration raises $R^2$ to $0.36$, and changing to the task-OOD split raises it to $0.43$. These sequential comparisons change one setup dimension at a time but do not form a complete factorial capacity-by-camera study, so we use them only to motivate the canonical two-camera DIFF configuration rather than to claim a general scaling rule.

\section{Temporal Cosine: A Diagnostic for the Auxiliary's Mechanism}
\label{app:cosine}

To diagnose how the inverse-dynamics loss modifies the trunk's representations, we measure the cosine similarity between mean-pooled features at temporal offset $k$ frames: $\mathrm{cos}(\bar{f}_t, \bar{f}_{t+k})$. A high cosine indicates that the features change slowly over time (\emph{temporally collapsed}); a low cosine indicates that the features change rapidly. Table~\ref{tab:cosine} reports cosine and action $R^2$ across the $\lambda$ sweep on DIFF.

\begin{table}[h]
    \centering
    \small
    \begin{tabular}{l c c c}
    \toprule
    Variant & $\mathrm{cos}(t, t+16)$ & Action $R^2$ & Rollout PSNR \\
    \midrule
    DIFF baseline (no ID)              & $0.999$ & $+0.426$ & $19.18$ \\
    DIFF $+$ ID $\lambda{=}0.01$ & $0.994$ & $\textbf{+0.598}$ & $19.07$ \\
    DIFF $+$ ID $\lambda{=}0.05$       & $0.972$ & $+0.572$ & $18.91$ \\
    DIFF $+$ ID $\lambda{=}0.1$        & $0.948$ & $+0.436$ & $18.54$ \\
    DIFF $+$ ID $\lambda{=}0.3$        & $0.900$ & $+0.282$ & $17.40$ \\
    DIFF $+$ ID $\lambda{=}1.0$        & $0.877$ & $+0.264$ & $13.75$ \\
    DIFF $+$ ID $\lambda{=}10$         & $0.652$ & $+0.145$ & $10.62$ \\
    \bottomrule
    \end{tabular}
    \caption{\textbf{Temporal cosine decreases monotonically with $\lambda$; action $R^2$ does not.} The baseline DIFF trunk has cosine $0.999$ at $k{=}16$. Increasing $\lambda$ reduces cosine from $0.999$ to $0.65$, while action $R^2$ peaks at $\lambda{=}0.01$ (cosine $0.994$) and then falls.}
    \label{tab:cosine}
\end{table}

\paragraph{Reading the cosine--$R^2$ dissociation.} Stronger ID weighting is accompanied by lower temporal cosine, but cosine does not monotonically predict action $R^2$. The small reduction from $0.999$ to $0.994$ coincides with the peak action readout, whereas larger reductions coincide with lower $R^2$. Average temporal cosine alone is therefore insufficient as a mechanism diagnostic for action recovery.

\paragraph{Limitation.} Cosine is a within-DIFF diagnostic. Although frozen pretrained checkpoints provide no-ID references, their objectives, feature geometry, and pooling differ from DIFF, and they were not trained from scratch under the matched $\lambda$ sweep. Their cosine values would therefore not isolate the same intervention. The cross-architecture multiplier result does not depend on this diagnostic.

\section{Cross-Benchmark Transfer}
\label{app:cross_benchmark}

We extend the cross-architecture analysis to two additional simulator benchmarks under the same probe protocol: CALVIN~\citep{calvin} ($24$ long-horizon tasks across $4$ fixed table environments, $7$-DoF) and MetaWorld~\citep{metaworld} (ML45 split: $45$ training tasks, $5$ held-out, $4$-DoF, scripted demos). For each benchmark we apply the ID fine-tune to eight backbones and evaluate on its task-OOD held-out split with the same $3000$-step MLP probe used in the main paper; Table~\ref{tab:cross_benchmark} reports the headline results.

\begin{table}[h]
    \centering
    \small
    \setlength{\tabcolsep}{6pt}
    \begin{tabular}{l c c c}
    \toprule
    Backbone $+$ ID  & LIBERO & CALVIN & MetaWorld \\
    \midrule
    V-JEPA 2 ViT-L                  & $\mathbf{+0.84}$ & $+0.86$          & $\mathbf{+0.59}$ \\
    V-JEPA 2.1 ViT-B                & $+0.82$          & $\mathbf{+0.88}$ & $+0.40$ \\
    VideoMAE V1 ViT-L               & $+0.75$          & $+0.80$          & $+0.35$ \\
    Web-DINO ViT-L                  & $+0.16$          & $+0.81$          & $-0.44$ \\
    SigLIP 2 ViT-L                  & $+0.17$          & $+0.77$          & $-0.17$ \\
    LAPA                            & $+0.51$          & $+0.78$          & $-0.35$ \\
    SDXL VAE                        & $-0.41$          & $+0.09$          & $-0.54$ \\
    Cosmos-1 tokenizer              & $-0.29$          & $+0.39$          & $-0.31$ \\
    \bottomrule
    \end{tabular}
    \caption{\textbf{Cross-benchmark comparison on three held-out evaluations.} LIBERO uses the fixed four-task subset described in Section~\ref{sec:setup}; CALVIN and MetaWorld use their held-out splits. V-JEPA $+$ ID is the top family on LIBERO and MetaWorld (V-JEPA $2$ ViT-L) and on CALVIN (V-JEPA $2.1$ ViT-B at $0.88$, V-JEPA $2$ ViT-L close behind at $0.86$). The Spearman rank correlation between LIBERO and MetaWorld backbone rankings is $\rho = +0.88$. CALVIN's $\rho$ with LIBERO is $+0.83$ and with MetaWorld is $+0.67$; CALVIN reorders image-only SSL backbones from near-zero R² on LIBERO and negative R² on MetaWorld to $0.77$--$0.81$ R². DIFF and Dreamer 4 are LIBERO-trained from scratch and omitted.}
    \label{tab:cross_benchmark}
\end{table}

The pattern across LIBERO and MetaWorld is consistent: video-pretrained backbones top the ranking, image-only SSL backbones land far below, and reconstruction encoders are at or below zero. MetaWorld is the sharpest version of this---image-only SSL and LAPA have negative R² in the $4$-DoF, $50$-task regime---and the $\rho = +0.88$ rank correlation shows that a similar ordering recurs on that benchmark. Because MetaWorld lacks the three LIBERO rotation dimensions ($drx, dry, drz$), the observed V-JEPA margin there cannot be attributed to those dimensions alone.

CALVIN is the outlier. A plausible explanation is that its four fixed table environments and relatively consistent layouts let per-frame appearance carry more of the action signal than in LIBERO or MetaWorld. This interpretation is consistent with the image-only SSL catch-up, but the benchmark comparison changes task distribution and action space as well as environment diversity, so it does not isolate the cause.

\paragraph{Probe-budget control.} An alternative reading of the CALVIN catch-up is that LIBERO's small probe-train budget (400 episodes) under-estimates image-only SSL while CALVIN's larger budget surfaces hidden action-relevant features. We address this by sweeping the LIBERO probe-train size at $\{400, 800, 1600, 3200\}$ episodes on Web-DINO $+$ ID and SigLIP 2 $+$ ID. Web-DINO $+$ ID R² rises from $0.16$ (400 ep) to $0.37$ (3200 ep) and SigLIP 2 $+$ ID from $-0.02$ to $0.26$, an 8$\times$ data increase yielding $\sim$$0.2$ R² lift. The remaining gap to V-JEPA $+$ ID at LIBERO probe-train 400 ($0.84$) is still $0.47$ for Web-DINO and $0.58$ for SigLIP 2. The CALVIN gap collapse is not reproduced at any LIBERO probe budget tested, so the LIBERO V-JEPA dominance and the CALVIN image-SSL catch-up arise from different mechanisms.

\section{V-JEPA Per-Layer Surgical Probe}
\label{app:vjepa_perlayer}

To diagnose where action information appears in the trunk, we probe each of V-JEPA $2$ ViT-L's $24$ transformer blocks for action $R^2$ on the fixed LIBERO held-out subset, both on the frozen pretrained checkpoint and on the $30$k-step LIBERO $+$ ID checkpoint (Table~\ref{tab:vjepa_perlayer}). Block $0$ is just after patch embedding. Features are block outputs taken before the encoder's final LayerNorm, so block $23$ is close to, but not identical with, the feature used by our main-text extractors.

\begin{table}[h]
    \centering
    \small
    \begin{tabular}{rccr}
\toprule
Block & Frozen $R^2$ & + ID $R^2$ & $\Delta$ \\
\midrule
0 & 0.193 & 0.084 & $-0.109$ \\
1 & 0.315 & 0.304 & $-0.011$ \\
2 & 0.411 & 0.364 & $-0.047$ \\
3 & 0.426 & 0.416 & $-0.011$ \\
4 & 0.451 & 0.421 & $-0.031$ \\
5 & 0.452 & 0.422 & $-0.030$ \\
6 & 0.459 & 0.462 & $+0.003$ \\
7 & 0.456 & 0.475 & $+0.020$ \\
8 & 0.435 & 0.429 & $-0.006$ \\
9 & 0.454 & 0.434 & $-0.021$ \\
10 & 0.444 & 0.474 & $+0.030$ \\
11 & 0.488 & 0.509 & $+0.021$ \\
12 & 0.474 & 0.559 & $+0.084$ \\
13 & 0.485 & 0.558 & $+0.072$ \\
14$^\dagger$ & 0.514 & 0.615 & $+0.101$ \\
15$^\dagger$ & \textbf{0.516} & 0.595 & $+0.079$ \\
16$^\dagger$ & 0.512 & 0.625 & $+0.113$ \\
17$^\dagger$ & 0.502 & 0.643 & $+0.141$ \\
18$^\dagger$ & 0.516 & 0.632 & $+0.116$ \\
19 & 0.487 & 0.658 & $+0.171$ \\
20 & 0.460 & 0.708 & $+0.248$ \\
21 & 0.422 & \textbf{0.721} & $+0.300$ \\
22 & 0.394 & 0.712 & $+0.317$ \\
23 & 0.425 & 0.687 & $+0.263$ \\
\bottomrule
\end{tabular}

    \caption{\textbf{V-JEPA 2 ViT-L per-block action $R^2$, before and after inverse-dynamics fine-tuning.} Mean of $3$ probe seeds; features are block outputs taken before the encoder's final LayerNorm. $^\dagger$ marks the frozen plateau (blocks within one probe-seed SD of the frozen maximum); bold marks the maximum of each column. The frozen trunk plateaus over blocks $14$--$18$ and drops by $\sim 0.12$ $R^2$ by block $22$, recovering only slightly at the final block. After $30$k steps of inverse-dynamics fine-tuning the profile shifts upward and peaks at block $21$ rather than at the final block.}
    \label{tab:vjepa_perlayer}
\end{table}

The frozen profile rises through the early and middle blocks, plateaus over blocks $14$--$18$ (maximum $0.516$ at block $15$; all five blocks lie within one probe-seed standard deviation of the maximum), then decreases through blocks $18$--$22$ before partially recovering at the final block. This pattern is consistent with a pretraining-induced difference between middle and late-block features, but does not by itself identify the predictor as its cause.

ID fine-tuning lifts the final four blocks by $+0.25$ to $+0.32$ R², considerably more than blocks $12$--$18$ ($+0.07$ to $+0.14$). The maximum moves from the block $14$--$18$ plateau to block $21$ after ID, while the final block stays $0.034$ R² below that new peak. Thus the lift is not uniform across blocks.

The layer profile supplies an exploratory diagnostic for Section~\ref{sec:rotation}; isolating the pretraining objective's contribution would require matched re-pretraining. The $+$ID curve uses the same $30$k checkpoint as Table~\ref{tab:multiplier}, but its pre-LayerNorm block outputs and separate probe execution explain why block $23$ ($0.425/0.687$) does not equal the main extractor's Table~\ref{tab:multiplier} value ($0.40/0.84$).

\section{Probe Data Efficiency Control}
\label{app:probe_efficiency}

An alternative reading of the CALVIN catch-up in Appendix~\ref{app:cross_benchmark} is that image-only SSL backbones look weak on LIBERO not because their features are weak but because LIBERO's probe-train budget (400 episodes) is too small to extract their action signal. We test this by encoding LIBERO probe features once per backbone and re-probing at sizes $\{400, 800, 1600, 3200\}$, with the same probe head and fixed four-task test subset (Table~\ref{tab:probe_efficiency}).

\begin{table}[h]
    \centering
    \small
    \begin{tabular}{c c c c c}
    \toprule
    Probe size & Web-DINO frozen & Web-DINO $+$ ID & SigLIP 2 frozen & SigLIP 2 $+$ ID \\
    \midrule
    $400$  & $+0.04$ & $+0.16$ & $+0.00$ & $-0.02$ \\
    $800$  & $+0.21$ & $+0.29$ & $+0.08$ & $+0.18$ \\
    $1600$ & $+0.21$ & $+0.33$ & $+0.03$ & $+0.21$ \\
    $3200$ & $+0.24$ & $\mathbf{+0.37}$ & $+0.12$ & $\mathbf{+0.26}$ \\
    \bottomrule
    \end{tabular}
    \caption{\textbf{LIBERO probe-train size sweep on image-only SSL backbones.} An $8\times$ increase in probe-train data lifts Web-DINO $+$ ID from $0.16$ to $0.37$ and SigLIP 2 $+$ ID from $-0.02$ to $0.26$. The $400$-episode row comes from a separate encoding and probing pass (probe seeds $11$--$13$) and therefore differs from Table~\ref{tab:multiplier} (probe seeds $0$--$2$), most visibly for SigLIP 2 $+$ ID ($-0.02$ vs.\ $0.17$); budget trends are compared within this pass.}
    \label{tab:probe_efficiency}
\end{table}

The control rules out the probe-budget hypothesis as the full explanation. At an $8\times$ probe-train budget, Web-DINO $+$ ID on LIBERO reaches $0.37$, still $0.47$ below V-JEPA $+$ ID at the original $400$-episode budget ($0.84$), and still $0.43$ below the same backbone's CALVIN result ($0.81$). The image-SSL ceiling on LIBERO does rise with more probe data, but it does not reach either V-JEPA's LIBERO performance or its own CALVIN performance. The cross-benchmark pattern in Appendix~\ref{app:cross_benchmark} is therefore not reducible to probe-data scaling.

\section{Training Settings for the Main Multiplier Matrix}
\label{app:training_settings}

Table~\ref{tab:training_settings} records the per-backbone optimization settings behind Table~\ref{tab:multiplier}. All pretrained encoders use AdamW, weight decay $0.05$, a $2$k-step warmup followed by cosine decay, $\mathcal{L}_{\mathrm{primary}}=0$, $\lambda=1$, full-encoder fine-tuning, and no encoder EMA. Batch is reported as the effective batch across GPUs.

\begin{table}[h]
    \centering
    \small
    \begin{tabular}{l c c c}
    \toprule
    Backbone & Updates & Effective batch & Learning rate \\
    \midrule
    V-JEPA 2 ViT-L / V-JEPA 2.1 ViT-B & $30$k & $4$ & $5\times10^{-5}$ \\
    VideoMAE V1 ViT-L & $20$k & $2$ & $5\times10^{-5}$ \\
    Web-DINO ViT-L / SigLIP 2 ViT-L & $20$k & $8$ & $1\times10^{-4}$ \\
    LAPA-LAQ-OpenX & $30$k & $2$ & $5\times10^{-5}$ \\
    SDXL VAE & $20$k & $16$ & $1\times10^{-4}$ \\
    Cosmos-1 tokenizer & $20$k & $16$ & $5\times10^{-5}$ \\
    \bottomrule
    \end{tabular}
    \caption{\textbf{Optimization settings for pretrained-encoder $+$ID rows in Table~\ref{tab:multiplier}.} DIFF and Dreamer 4 are trained from scratch rather than fine-tuned under this schedule: canonical DIFF uses $100$k updates, batch $8$, lr $1.5\times10^{-4}$, $\lambda=0.05$, and diffusion EMA; the Dreamer 4 rows use the checkpoint-specific settings described in Appendix~\ref{app:backbone_catalog}.}
    \label{tab:training_settings}
\end{table}

\section{Detailed Cross-Architecture Multiplier Table}
\label{app:full_multiplier_table}

Table~\ref{tab:multiplier_detailed} reports the per-variant breakdown of Table~\ref{tab:multiplier} in the main text, with parameter counts and individual frozen / $+$ ID R² for every backbone we evaluated.

\begin{table}[h]
    \centering
    \small
    \setlength{\tabcolsep}{3pt}
    \begin{tabular}{l l c c c c}
    \toprule
    Pretraining family & Backbone & Params & Frozen $R^2$ & $+$ ID $R^2$ & $\Delta$ \\
    \midrule
    \multirow{2}{*}{Video $+$ JEPA pred.}
        & V-JEPA 2 ViT-L                   & $304$M & $0.40$  & $\textbf{0.84}$           & $\textbf{+0.44}$ \\
        & V-JEPA 2.1 ViT-B (distilled)     & $87$M  & $0.44$  & $0.82$                    & $+0.38$ \\
    \midrule
    Video $+$ pixel MAE
        & VideoMAE V1 ViT-L                & $304$M & $0.46$  & $0.75$                    & $+0.29$ \\
    \midrule
    \multirow{2}{*}{Image-only SSL}
        & Web-DINO ViT-L                   & $304$M & $-0.01$ & $0.16$                    & $+0.17$ \\
        & SigLIP 2 ViT-L                   & $316$M & $0.05$  & $0.17$                    & $+0.12$ \\
    \midrule
    Latent action quant.
        & LAPA-LAQ-OpenX                   & $344$M & $0.41$  & $0.51$                    & $+0.10$ \\
    \midrule
    Pixel generation
        & DIFF (pixel diffusion)           & $91$M  & $0.43$  & $0.57$                    & $+0.15$ \\
    \midrule
    \multirow{2}{*}{Pixel reconstruction}
        & SDXL VAE                         & $34$M  & $-0.55$ & $-0.41$                   & $+0.14$ \\
        & Cosmos-$1$ tokenizer             & $34$M  & $-0.36$ & $-0.29$                   & $+0.07$ \\
    \midrule
    \multirow{3}{*}{Shortcut-forcing}
        & Dreamer 4 (scratch $64$M)        & $64$M  & $-0.04$ & $-0.04$                   & $0.00$ \\
        & Dreamer 4 (DMControl init $64$M) & $64$M  & $-0.04$ & $-0.04$                   & $0.00$ \\
        & Dreamer 4 (scratch $276$M)       & $276$M & n/a     & $-0.04$                   & n/a \\
    \bottomrule
    \end{tabular}
    \caption{\textbf{Detailed per-variant multiplier table.} Backbones grouped by pretraining objective. ``n/a'' indicates a configuration not run: the $276$M Dreamer 4 row is an ID-only variant. The DMControl row reports the agent-token stream for both frozen and $+$ID configurations. The Cosmos-$1$ tokenizer row matches the exact encoder used by~\citet{semantic_wm_2026}.}
    \label{tab:multiplier_detailed}
\end{table}

\section{Full Visual-Perturbation Robustness Aggregates}
\label{app:full_robustness}

\begin{table}[h]
    \centering
    \small
    \setlength{\tabcolsep}{4pt}
    \begin{tabular}{l l r r r r r}
    \toprule
    Family & Backbone & clean & noise $0.10$ & noise $0.15$ & blur $11$ & worst \\
    \midrule\midrule
    \multicolumn{7}{l}{\emph{$+$ ID variants}} \\
    \midrule
    \multirow{2}{*}{Video $+$ JEPA pred.}
        & V-JEPA 2 ViT-L           & $\textbf{0.84}$ & $\textbf{0.75}$ & $\textbf{0.66}$ & $0.56$          & $0.56$ \\
        & V-JEPA 2.1 ViT-B         & $0.82$          & $0.46$          & $0.01$          & $\textbf{0.70}$ & $0.01$ \\
    \midrule
    Video $+$ Pixel MAE          & VideoMAE V1 ViT-L        & $0.75$ & $0.39$ & $0.11$ & $0.55$ & $0.11$ \\
    \midrule
    \multirow{2}{*}{Image SSL}
        & Web-DINO ViT-L           & $0.16$ & $0.13$ & $0.08$ & $-0.05$ & $-0.05$ \\
        & SigLIP 2 ViT-L           & $0.17$ & $0.07$ & $-0.09$ & $-0.40$ & $-0.40$ \\
    \midrule
    Latent action quant.             & LAPA-LAQ-OpenX           & $0.51$ & $0.14$ & $-0.19$ & $0.50$ & $-0.19$ \\
    \midrule
    Pixel diffusion              & DIFF                     & $0.57$ & $0.57$ & $0.57$ & $0.52$ & $0.46$ \\
    \midrule
    \multirow{2}{*}{Pixel reconstruction}
        & SDXL VAE                 & $-0.41$ & $-1.69$ & $-2.20$ & $-2.02$ & $-2.20$ \\
        & Cosmos-1 tokenizer       & $-0.29$ & $-0.66$ & $-0.86$ & $-1.09$ & $-1.09$ \\
    \midrule
    Shortcut-forcing dyn.        & Dreamer 4                & $-0.04$ & $-0.04$ & $-0.04$ & $-0.04$ & $-0.04$ \\
    \midrule
    \multicolumn{7}{l}{\emph{Frozen baselines}} \\
    \midrule
    \multirow{2}{*}{Video $+$ JEPA pred.}
        & V-JEPA 2 ViT-L           & $0.40$ & $-0.82$ & $-1.18$ & $\mathbf{-4.66}$ & $-4.66$ \\
        & V-JEPA 2.1 ViT-B         & $0.44$ & $\mathbf{-4.31}$ & $\mathbf{-5.36}$ & $-0.86$ & $-5.36$ \\
    \midrule
    Video $+$ Pixel MAE          & VideoMAE V1 ViT-L        & $0.46$ & $-2.02$ & $-2.54$ & $-1.10$ & $-2.54$ \\
    \midrule
    \multirow{2}{*}{Image SSL}
        & Web-DINO ViT-L           & $-0.01$ & $-3.40$ & $-4.19$ & $-1.60$ & $-4.19$ \\
        & SigLIP 2 ViT-L           & $0.05$ & $-0.63$ & $-0.62$ & $-1.75$ & $-1.75$ \\
    \midrule
    Latent action quant.             & LAPA-LAQ-OpenX           & $0.41$ & $-0.24$ & $-1.14$ & $0.38$ & $-1.14$ \\
    \midrule
    Pixel diffusion              & DIFF                     & $0.43$ & $0.43$ & $0.44$ & $0.42$ & $0.24$ \\
    \midrule
    \multirow{2}{*}{Pixel reconstruction}
        & SDXL VAE                 & $-0.55$ & $-1.57$ & $-1.96$ & $-1.52$ & $-1.96$ \\
        & Cosmos-1 tokenizer       & $-0.36$ & $-0.59$ & $-0.80$ & $-0.94$ & $-0.94$ \\
    \midrule
    Shortcut-forcing dyn.        & Dreamer 4                & $-0.04$ & $-0.04$ & $-0.04$ & $-0.04$ & $-0.04$ \\
    \bottomrule
    \end{tabular}
    \caption{\textbf{Robustness to visual perturbation on the fixed four-task held-out subset, with full per-condition results for Table~\ref{tab:perturb}.} The probe is trained on clean features and evaluated on perturbed inputs; ``Worst'' is the minimum across all seven conditions. Frozen DIFF is in-distribution for LIBERO and floors at clean $0.43$ across noise, while Dreamer 4 floors at $-0.04$ throughout. Per-family aggregates are in Table~\ref{tab:robustness_summary}.}
    \label{tab:perturb_full}
\end{table}

Per-family averages (Table~\ref{tab:robustness_summary}) summarize the per-perturbation breakdown beyond the headline subset reported in Table~\ref{tab:perturb}. The paired rows in Table~\ref{tab:perturb} show that $+$ID variants are generally more robust for the video-pretrained and image-SSL checkpoints; the reconstruction-aligned checkpoints are exceptions, and Dreamer 4 remains at its floor.

\begin{table}[h]
    \centering
    \small
    \setlength{\tabcolsep}{4pt}
    \begin{tabular}{l c c c c c}
    \toprule
    Family & $N$ & clean & noise $0.10$ & noise $0.15$ & blur $11$ \\
    \midrule
    V-JEPA 2                          & $4$  & $+0.48$ & $-0.24$ & $-0.57$ & $-1.05$ \\
    V-JEPA 2.1                        & $2$  & $+0.63$ & $-1.92$ & $-2.68$ & $-0.08$ \\
    VideoMAE                          & $2$  & $+0.60$ & $-0.81$ & $-1.22$ & $-0.28$ \\
    LAPA                              & $2$  & $+0.46$ & $-0.05$ & $-0.67$ & $+0.44$ \\
    DIFF                              & $55$ & $+0.16$ & $+0.14$ & $+0.11$ & $-0.02$ \\
    Web-DINO                          & $2$  & $+0.08$ & $-1.64$ & $-2.05$ & $-0.83$ \\
    SigLIP 2                          & $2$  & $+0.11$ & $-0.28$ & $-0.35$ & $-1.08$ \\
    Cosmos-1 tokenizer                & $2$  & $-0.33$ & $-0.63$ & $-0.83$ & $-1.02$ \\
    SDXL VAE                          & $2$  & $-0.48$ & $-1.63$ & $-2.08$ & $-1.77$ \\
    Dreamer 4                         & $10$ & $-0.07$ & $-0.13$ & $-0.16$ & $-0.15$ \\
    \bottomrule
    \end{tabular}
    \caption{\textbf{Per-family robustness aggregates.} $N$ is the variant count in the family (frozen and $+$ ID variants mixed; the large DIFF count comes from its capacity, $\lambda$, and head-form sweeps documented in Appendix~\ref{app:aux_ablations} and Appendix~\ref{app:diff_findings}). These mixed family means should not be interpreted as paired ID effects: $+$ID variants are generally more robust for the video-pretrained and image-SSL checkpoints, reconstruction-aligned checkpoints are exceptions, and Dreamer 4 remains at its floor.}
    \label{tab:robustness_summary}
\end{table}

\section{ID Supervision Sample-Budget Sweep}
\label{app:id_sparsity}

\begin{figure}[h]
    \centering
    \includegraphics[width=0.85\linewidth]{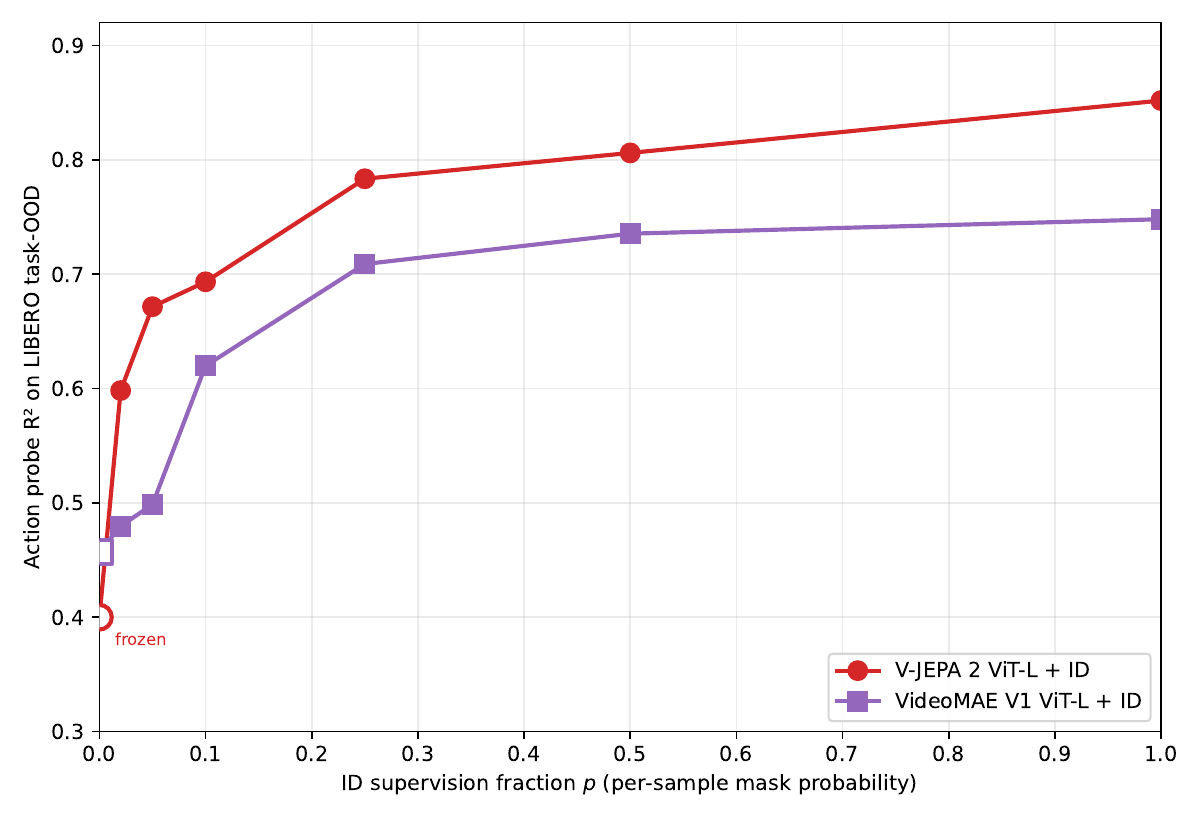}
    \caption{\textbf{ID supervision sample-budget sweep on V-JEPA $2$ ViT-L and VideoMAE V$1$ ViT-L.} $x$-axis: fraction $p$ of mini-batch sample slots receiving ID gradient (per-sample mask probability). $y$-axis: action-probe $R^2$ on the fixed four-task LIBERO held-out subset, mean of $3$ probe seeds. The $p{=}0$ points are frozen references from Table~\ref{tab:multiplier}; $p{>}0$ entries come from a separate encoding and probing pass. V-JEPA at $p{=}0.02$ is approximately $0.20$ above that reference, but this cross-pass difference is descriptive. At $p{=}0.02$, about $1600$ sample-slot assignments are expected to receive ID gradient over $20$k$\times$$4$ slots.}
    \label{fig:sparsity_sweep}
\end{figure}

To probe the mechanism by which the inverse-dynamics loss interacts with V-JEPA's pretrained features, we vary the fraction $p$ of mini-batch samples that receive ID gradient. For each sample in a mini-batch of size $B$, we draw a Bernoulli($p$) mask; samples with mask$=0$ contribute zero loss (their gradient is skipped). Batch size ($4$ for V-JEPA, $2$ for VideoMAE), learning rate ($5 \times 10^{-5}$), $\lambda{=}1.0$, no encoder EMA, and the probe architecture match Table~\ref{tab:multiplier}. Every sweep run uses $20$k fine-tune steps, matching the VideoMAE row; the V-JEPA row of Table~\ref{tab:multiplier} uses $30$k steps. At $p{=}0$ no encoder gradient is produced, recovering the frozen baseline; at $p{=}1$ every sample contributes ID loss. We sweep $p \in \{0.02, 0.05, 0.10, 0.25, 0.50, 1.00\}$ on both V-JEPA $2$ ViT-L and VideoMAE V$1$ ViT-L (Figure~\ref{fig:sparsity_sweep}). The sweep runs were probed in a separate encoding pass with different clip sampling and probe seeds, so the comparison includes evaluation variance; the $p{=}0$ entries in Table~\ref{tab:sparsity_sweep} are the Table~\ref{tab:multiplier} frozen values.

\begin{table}[h]
    \centering
    \small
    \setlength{\tabcolsep}{6pt}
    \begin{tabular}{c c c c c c c c}
    \toprule
    & \multicolumn{7}{c}{ID supervision fraction $p$} \\
    \cmidrule(lr){2-8}
    Backbone & $0.00$ & $0.02$ & $0.05$ & $0.10$ & $0.25$ & $0.50$ & $1.00$ \\
    \midrule
    V-JEPA $2$ ViT-L $+$ ID  & $0.400$ & $\mathbf{0.598}$ & $0.672$ & $0.693$ & $0.783$ & $0.806$ & $\mathbf{0.852}$ \\
    VideoMAE V$1$ ViT-L $+$ ID & $0.457$ & $0.480$ & $0.499$ & $0.620$ & $0.709$ & $0.735$ & $0.748$ \\
    \bottomrule
    \end{tabular}
    \caption{\textbf{Aggregate action $R^2$ across the sparsity sweep.} Mean of $3$ probe seeds. Within the separate sweep pass, V-JEPA at $p{=}0.02$ ($0.598$) exceeds VideoMAE at $p\leq0.05$; this pattern is consistent with greater sample efficiency for the evaluated V-JEPA checkpoint.}
    \label{tab:sparsity_sweep}
\end{table}

\paragraph{Findings.} Within the $p{>}0$ sweep pass, four patterns are visible. (i) V-JEPA rises quickly: $R^2$ is $0.598$ at $p{=}0.02$ and $0.693$ at $p{=}0.10$. Relative to the Table~\ref{tab:multiplier} frozen reference, these are approximate cross-pass differences rather than controlled within-pass lifts. (ii) VideoMAE changes little from $p{=}0.02$ to $0.05$ ($0.480\to0.499$) and rises more strongly at $p\geq0.10$. (iii) Translation ($dx, dy, dz$) reaches $\geq0.94$ by $p{=}0.05$ on both backbones, while rotation ($drx, dry, drz$) requires denser supervision in the reported runs. (iv) V-JEPA exceeds VideoMAE at each matched nonzero $p$; at $p{=}0.02$ the gap is $0.12$. These observations are consistent with greater sample efficiency for this V-JEPA checkpoint, but the sweep does not establish that ID merely re-aligns pre-existing signal or isolate the pretraining objective as the cause.

\paragraph{Implementation detail.} The per-sample masking is implemented as \texttt{mask = torch.rand(B) < p}; the loss is \texttt{(mask.float() * per\_sample\_id\_loss).sum() / mask.sum().clamp(min=1)}, with the step skipped entirely if \texttt{mask.sum() == 0}. At $p{=}0.02$, $B{=}4$, this gives expected $0.08$ active sample slots per step, so the majority of mini-batches contribute zero gradient---approximately $1600$ sample-slot assignments are selected across the full $20$k$\times$$4$ training run (not necessarily $1600$ unique samples).

\section{Linear-Probe Action Subspace Visualization}
\label{app:probe_subspace}

To visualize the effect of ID supervision at the episode-trajectory level, we project features from the fixed four-task LIBERO held-out subset into a canonical 2D action subspace. This visualization is exploratory and is not used as evidence for the main-text claims. For each backbone, we fit a Ridge linear regressor $\hat{a} = X W$ (regularization $\alpha{=}1.0$) on training-set features, then take the top-2 left-singular vectors of $W \in \mathbb{R}^{D \times 7}$ as feature-space directions. Per-episode feature sequences are projected onto these two directions and plotted as time-ordered trajectories.

\begin{figure}[h]
    \centering
    \includegraphics[width=\linewidth]{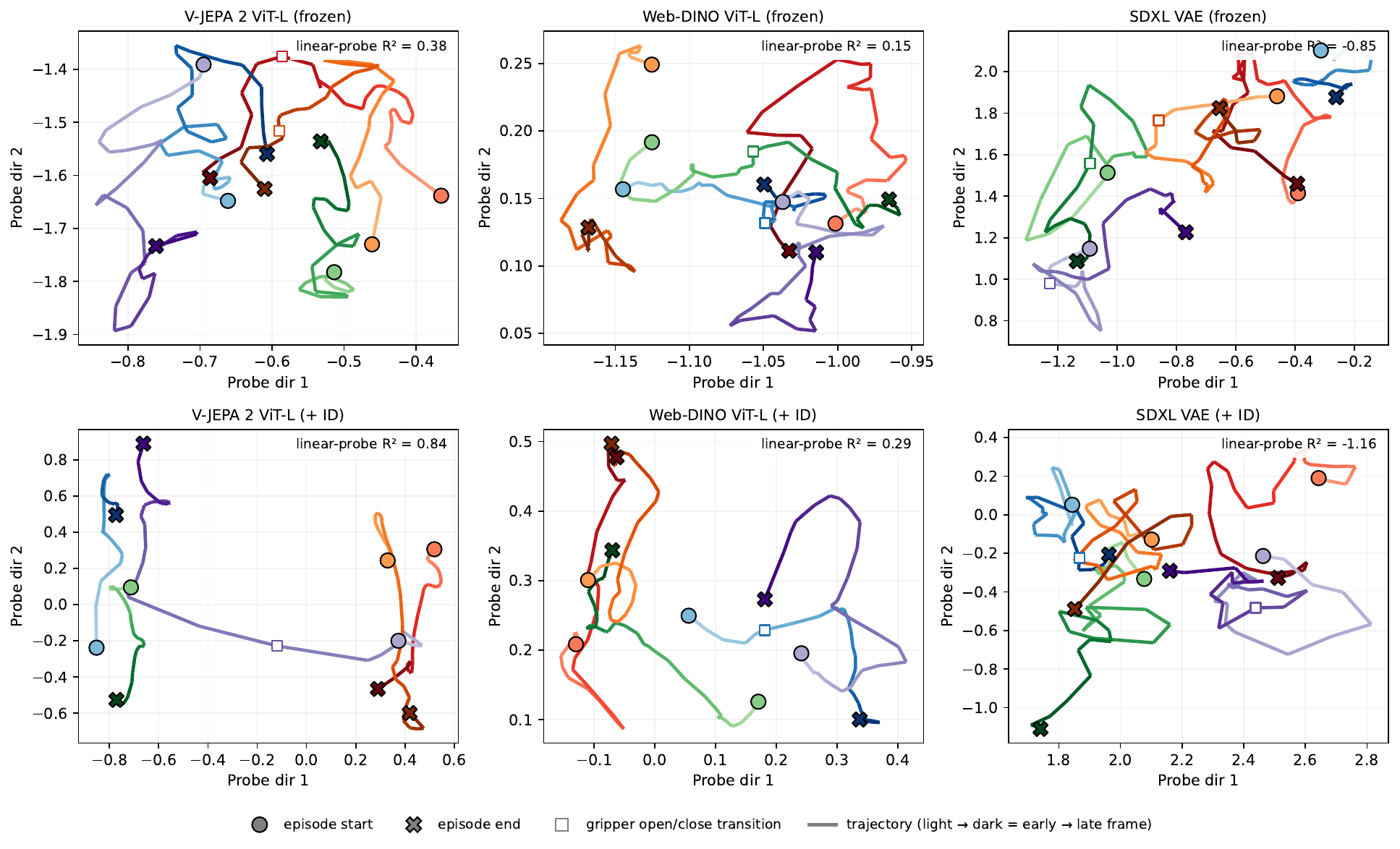}
    \caption{\textbf{Linear-probe action subspace trajectories on the fixed four-task LIBERO held-out subset.} Top row: frozen backbones. Bottom row: $+$ ID fine-tunes. Each colored curve is one episode ($5$ episodes shown, fixed random seed across panels; $3$-frame moving average smoothing applied). $\bigcirc$ = episode start, $\times$ = episode end, $\square$ = gripper open/close transition. Color saturation encodes time within an episode (light $\to$ dark $=$ early $\to$ late frame). The linear-probe $R^2$ shown in each panel is computed on the same 2D subspace as the trajectories. The contrast across columns visualizes the multiplier effect documented in Table~\ref{tab:multiplier}: V-JEPA $+$ ID (bottom-left, $R^2 = 0.84$) produces episode-distinct trajectories spanning the entire 2D subspace; Web-DINO $+$ ID (bottom-middle, $R^2 = 0.29$) shows partial improvement with trajectories still compressed; SDXL VAE $+$ ID (bottom-right, $R^2 = -1.16$) regresses relative to its frozen baseline ($R^2 = -0.85$).}
    \label{fig:probe_subspace}
\end{figure}

\paragraph{Linear vs.\ MLP probe.} The $R^2$ values shown in Figure~\ref{fig:probe_subspace} are linear-probe (Ridge regression) accuracy on the 2D top-SVD subspace of the linear regressor weight matrix; the corresponding MLP-probe values reported in Table~\ref{tab:multiplier} are computed in the full feature space and differ accordingly (V-JEPA: linear $0.84$ vs.\ MLP $0.84$; V-JEPA frozen: linear $0.38$ vs.\ MLP $0.40$; Web-DINO $+$ ID: linear $0.29$ vs.\ MLP $0.16$; SDXL VAE $+$ ID: linear $-1.16$ vs.\ MLP $-0.41$). Backbone ordering on the V-JEPA $\to$ Web-DINO $\to$ SDXL VAE axis is preserved across both probes; the linear-probe subspace is used here because it admits a closed-form top-$2$ SVD, providing a canonical 2D visualization comparable across backbones.

\paragraph{Reading the figure.} Three patterns are visible. \emph{V-JEPA frozen $\to$ $+$ ID}: trajectory spread expands from a narrow horizontal band (probe dir 2 range $\sim 0.5$) to a 2D fan (range $\sim 1.5$); episode endpoints separate into distinct clusters at the panel corners, indicating that different episodes terminate in different action-relevant states. \emph{Web-DINO frozen $\to$ $+$ ID}: trajectories sharpen modestly with the ID loss but most episode trajectories remain mutually overlapping---consistent with the family's negative rotation $R^2$ in Figure~\ref{fig:perdim}. \emph{SDXL VAE frozen $\to$ $+$ ID}: the frozen panel shows trajectories scattered across the subspace; the $+$ ID panel shows trajectories \emph{compressed} into a denser cluster around probe dir 2 $\in [-0.5, 0]$ while linear-probe $R^2$ drops from $-0.85$ to $-1.16$, consistent with pixel-reconstruction-aligned features moving further from the action manifold under ID supervision rather than toward it. The episode-level visualization thus reproduces the same family ordering as the aggregate per-dimension and robustness tables.

\section{Backbone Catalog}
\label{app:backbone_catalog}

Detailed description of the ten reported backbone entries spanning seven objective families.

\paragraph{V-JEPA 2 (ViT-L, ViT-B).} A self-supervised video encoder trained with feature-level masked latent prediction~\citep{vjepa2}: the model is given a video clip with most spatio-temporal patches masked and learns to predict the encoder's own features at the masked positions, via a target-encoder $+$ predictor architecture. We use the public V-JEPA $2$ ViT-L checkpoint ($304$M parameters) and the distilled V-JEPA $2.1$ ViT-B checkpoint ($87$M parameters). Pretraining is on internet-scale video. The output token sequence has spatio-temporal structure; we average-pool across patch tokens within each frame and stack across frames to form the per-step feature vector.

\paragraph{VideoMAE V1 ViT-L.} A self-supervised video encoder trained with pixel-space masked autoencoding~\citep{videomae}: most spatio-temporal patches of a video clip are masked and a decoder reconstructs raw RGB pixels at the masked positions. We use the public VideoMAE V1 ViT-L checkpoint ($304$M parameters) pretrained on Kinetics-$400$. We discard the decoder at evaluation and use the encoder's patch-token features identically to V-JEPA.

\paragraph{Web-DINO ViT-L.} A self-supervised image encoder trained with masked contrastive prediction on internet images~\citep{webssl}; conceptually a successor to DINOv2 that scales the SSL recipe to web-scale data. We use the public Web-DINO ViT-L checkpoint ($304$M parameters). Per-frame patch-token features are pooled the same way as the video encoders; temporal context, when used, comes only from stacking consecutive per-frame features.

\paragraph{SigLIP 2 ViT-L.} A vision-language contrastive encoder trained with sigmoid-loss image-text pairs on a large web-scale image-text corpus~\citep{siglip2}. We use the public SigLIP 2 ViT-L checkpoint ($316$M parameters) and extract patch-token features from the vision side. Language supervision is the axis that distinguishes SigLIP 2 from Web-DINO in our taxonomy.

\paragraph{LAPA-LAQ-OpenX.} A latent-action quantizer pretrained on the Open X-Embodiment robot manipulation corpus~\citep{lapa}. The pretraining objective is to map consecutive video frames to a discrete latent-action code, learning a representation aligned with manipulation dynamics. We use the public LAPA-LAQ checkpoint ($344$M parameters). On the fixed subset, LAPA reaches moderate $R^2$ ($0.51$ $+$ ID) but remains below V-JEPA ($0.84$); this checkpoint comparison does not isolate the effect of action-explicit pretraining.

\paragraph{DIFF.} A pixel-diffusion world model trained from scratch on LIBERO at $34$--$91$M parameters. The canonical $91$M configuration is a $2$-camera ViT-style diffusion transformer (patch size $16$, width $768$, $12$ blocks and $12$ attention heads) predicting the next frame from a window of past frames. It is trained without action input on the $104$ training tasks ($400$ episodes); action conditioning hurts representation in the within-DIFF study of Appendix~\ref{app:diff_findings}. Feature extraction is from the diffusion trunk at zero diffusion timestep.

\paragraph{SDXL VAE and Cosmos-1 image tokenizer.} Two pixel-reconstruction-aligned autoencoders. The SDXL VAE~\citep{sdxlvae} is the publicly released image VAE from the SDXL diffusion stack ($34$M parameters), pretrained on LAION-$5$B with an MSE $+$ LPIPS $+$ KL reconstruction objective. The Cosmos-1 image tokenizer~\citep{cosmos1} is the continuous image tokenizer used in NVIDIA's Cosmos-1 world-model stack ($34$M parameters), pretrained on the Cosmos-1 video corpus. The Cosmos-1 checkpoint we use is the exact checkpoint employed by~\citet{semantic_wm_2026}, enabling direct comparison with their results. Both are encoder-only backbones in our study; we extract patch-token features from the encoder and discard the decoder.

\paragraph{Dreamer 4.} An unofficial reproduction of the shortcut-forcing dynamics architecture introduced by~\citet{dreamerv4}. Since the original implementation is not publicly released, we build on the unofficial PyTorch implementation\footnote{\url{https://github.com/nicklashansen/dreamer4}} and use its DMControl-pretrained checkpoint for the initialization experiment. We train $64$M- and $276$M-parameter variants on LIBERO and probe both agent-token and spatial-token feature streams. The agent-token stream is the canonical model state (a sequence of $\sim 100$ latent tokens combining stochastic and deterministic state); the spatial-token stream is the ViT trunk's patch features before agent-token compression. Baseline world-model quality is not independently verified, so all Dreamer 4 results are exploratory.

\section{Matched-Control and CAIP Details}
\label{app:matched}

\paragraph{Training.} Starting from identical V-JEPA $2$ ViT-L, VideoMAE V1 ViT-L, and Web-DINO ViT-L checkpoints, we train true ID, within-task action-shuffled ID, and forward dynamics (FD) for $20$k optimizer updates with two training seeds each. All runs use AdamW with learning rate $5{\times}10^{-5}$, a cosine schedule with $2$k warmup steps, weight decay $0.05$, $\mathcal{L}_{\text{primary}}{=}0$, auxiliary weight $1$, and no encoder EMA. Micro-batches are $4/2/4$ for V-JEPA $2$/VideoMAE/Web-DINO, with gradient accumulation to effective batch $16$. Initialization, clips, schedule, augmentations, and checkpointing are matched across the three objectives within each backbone. Action-shuffled ID permutes actions among transitions of the same task, so the action marginal and the data are unchanged but each transition is paired with the wrong action. FD predicts $\mathrm{sg}(z_{t+1})$ from $(z_t, a_t)$. After training we discard the training heads and fit three fresh $[D \to 256 \to 128 \to 7]$ probes.

\paragraph{Evaluation.} The absolute columns of Table~\ref{tab:matched} use the fixed $200$ held-out episodes and average two training seeds $\times$ three probe seeds (frozen: three probe seeds). $\Delta_{26}$ recomputes pooled per-dimension $R^2$ over all $26$ held-out tasks ($50$ episodes per task) and reports a $95\%$ paired held-out-task cluster-bootstrap CI ($10$k replicates), conditional on the two trained checkpoints and three probe seeds. Table~\ref{tab:matched_worst} reports the same True$-$shuffled contrast under the worst visual corruption of Section~\ref{sec:robustness}.

\begin{table}[h]
    \centering
    \small
    \begin{tabular}{l c c}
    \toprule
    Encoder & Clean $\Delta_{26}$ & Worst-corruption $\Delta_{26}$ \\
    \midrule
    V-JEPA 2 ViT-L    & $0.836$ [$0.752$, $0.952$] & $0.662$ [$0.595$, $0.754$] \\
    VideoMAE V1 ViT-L & $0.578$ [$0.518$, $0.658$] & $0.686$ [$0.638$, $0.763$] \\
    Web-DINO ViT-L    & $0.097$ [$0.026$, $0.156$] & $-0.026$ [$-0.108$, $0.036$] \\
    \bottomrule
    \end{tabular}
    \caption{\textbf{True $-$ action-shuffled ID gap, clean and under the worst corruption.} Pooled per-dimension $R^2$ over the $26$ held-out tasks with task-cluster bootstrap intervals.}
    \label{tab:matched_worst}
\end{table}

\paragraph{CAIP and evaluation coverage.} We use the official CAIP ViT-L checkpoint~\citep{caip} with its official preprocessing and mean-pool its visual patch tokens before the probe. Table~\ref{tab:caip_coverage} reports a separate encoding pass under the Table~\ref{tab:multiplier} probe settings. On the fixed $200$-episode subset, visual-only CAIP reaches $R^2=0.019$, versus $0.449$ for V-JEPA $2$, $0.472$ for VideoMAE, $0.140$ for Web-DINO, and $-0.024$ for the constant predictor. Over all $26$ held-out tasks, absolute frozen performance decreases substantially; visual-only CAIP is $-1.446$ [$-1.961$, $-1.055$] relative to the constant predictor. Supplying the LIBERO task description to CAIP does not improve action recovery under this probe.

\begin{table*}[t]
    \centering
    \small
    \resizebox{\textwidth}{!}{\begin{tabular}{lccc}
\toprule
 & $R^2$ (200 ep) & $R^2$ (26 tasks) [95\% CI] & $R^2 -$ constant (26 tasks) [95\% CI] \\
\midrule
CAIP ViT-L, visual only & 0.019 & $-$1.480 [$-$1.998, $-$1.093] & $-$1.446 [$-$1.961, $-$1.055] \\
V-JEPA 2 ViT-L (frozen) & 0.449 & $-$0.006 [$-$0.146, 0.099] & 0.028 [$-$0.107, 0.140] \\
VideoMAE V1 ViT-L (frozen) & 0.472 & $-$0.186 [$-$0.405, $-$0.005] & $-$0.151 [$-$0.365, 0.035] \\
Web-DINO ViT-L (frozen) & 0.140 & $-$1.432 [$-$2.031, $-$0.979] & $-$1.398 [$-$1.991, $-$0.941] \\
Constant predictor & $-$0.024 & -- & -- \\
\midrule
CAIP ViT-L, text-conditioned$^{*}$ & $-$2.762 & $-$3.950 [$-$6.196, $-$2.471] & $-$3.916 [$-$6.152, $-$2.432] \\
\bottomrule
\end{tabular}
}
    \caption{\textbf{Action-centric comparator and frozen-probe coverage check.} Frozen features come from a separate encoding pass with the Table~\ref{tab:multiplier} probe settings (MLP $D\!\to\!256\!\to\!128\!\to\!7$, AdamW, lr $3\times10^{-4}$, batch $512$, $3000$ steps, $3$ probe seeds; probe training uses the same $400$ episodes). The separate pass explains small differences from Table~\ref{tab:multiplier}. ``200 ep'' is the fixed four-task test pool; ``26 tasks'' uses all $1{,}300$ held-out episodes with a $10$k-replicate held-out-task cluster-bootstrap $95\%$ CI. The text-conditioned CAIP row receives the LIBERO task description and is reported separately from the vision-only ranking.}
    \label{tab:caip_coverage}
\end{table*}

\section{Closed-Loop Target-Swap Details}
\label{app:closedloop}

\paragraph{Policy training.} All policies start from the $\pi_{0.5}$ base checkpoint~\citep{pi05} and are fine-tuned on all five LIBERO suites (LIBERO-90, -10, -Goal, -Object, -Spatial; $90\%$ of demonstrations) for $10$k steps with batch size $32$, peak learning rate $1.2{\times}10^{-4}$, $200$ warmup steps, action horizon $16$, and bf16 precision. The auxiliary loss is applied from the first step. The target encoder is frozen and is used only to compute training targets; it is not needed at inference.

\begin{figure}[htb]
    \centering
    \includegraphics[width=\linewidth]{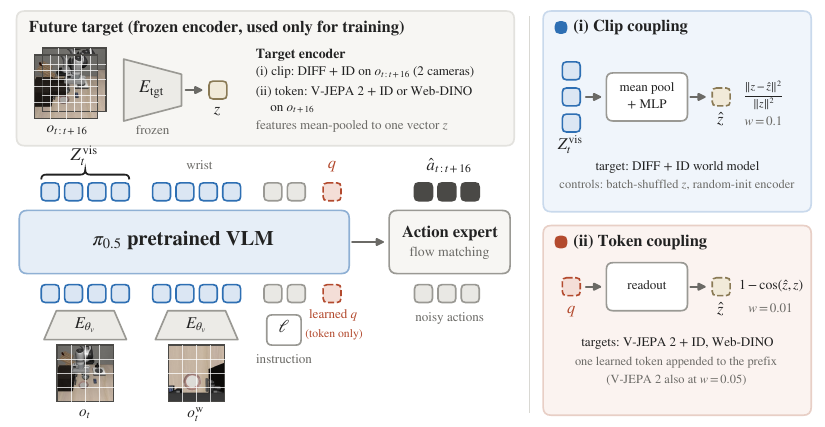}
    \caption{\textbf{Closed-loop target-swap setup.} $\pi_{0.5}$ is fine-tuned with its flow-matching action loss plus an auxiliary loss toward $z$, the future latent of a frozen target encoder that is used only during training. (i)~Clip coupling: the main-camera prefix tokens $Z_t^{\mathrm{vis}}$ are mean-pooled and mapped by an MLP to $\hat z$, which is trained toward the DIFF $+$ ID encoding of the two-camera clip $o_{t:t+16}$ with a normalized $L_2$ loss (weight $0.1$); the batch-shuffled and random-init controls change only the target. (ii)~Token coupling: a learned token $q$ is appended to the prefix, and a readout of its output is trained toward the mean-pooled V-JEPA $2$ $+$ ID or Web-DINO feature of $o_{t+16}$ with a cosine loss (weight $0.01$).}
    \label{fig:closedloop_method}
\end{figure}

\paragraph{Clip coupling.} An MLP (hidden width $2048$) on $\pi_{0.5}$'s pooled main-camera prefix tokens predicts the frozen DIFF $+$ ID encoding of the two-camera clip $o_{t}, \dots, o_{t+16}$ (final-step tokens, mean-pooled), with $\mathcal{L}_{\text{aux}} = \|z - \hat z\|_2^2 / \|z\|_2^2$ and weight $0.1$ (Figure~\ref{fig:closedloop_method}). The saved target configuration and run path match the Table~\ref{tab:multiplier} DIFF $+$ ID configuration (probe $R^2=0.5724$), although we did not verify byte-level weight identity. The batch-shuffled control permutes targets across the mini-batch, so each sample is trained toward another sample's target. The random-init control uses the same DIFF architecture without loading trained weights, which keeps a deterministic target of the correct future clip but removes learned structure.

\paragraph{Token coupling.} A learned token is appended to $\pi_{0.5}$'s prefix, and a readout of its output predicts the target encoder's mean-pooled feature of $o_{t+16}$ under a cosine loss (weight $0.01$; we also report $0.05$ for V-JEPA $2$). Targets come from our LIBERO ID-fine-tuned V-JEPA $2$ ViT-L checkpoint or the public frozen Web-DINO ViT-L checkpoint.

\paragraph{Evaluation and full results.} LIBERO uses $10$ tasks $\times$ $10$ trials per suite. LIBERO-PRO~\citep{libero_pro} applies five shifts (language, object position, object, task, environment) to each of the four suites, again with $10$ tasks $\times$ $10$ trials. Table~\ref{tab:closedloop_full} reports every shift. The position-shift gain is specific to the long-horizon suite: averaged over suites, the DIFF $+$ ID target's position-shift success ($47.5$) is within one point of the baseline ($46.8$). A second policy seed using that target is lower on most shifts ($65.1$ overall), and V-JEPA $2$ at weight $0.05$ loses the position-shift gain ($8$), so these closed-loop effects are sensitive to policy seed and auxiliary weight.

\begin{table}[h]
    \centering
    \footnotesize
    \setlength{\tabcolsep}{3pt}
    \begin{tabular}{l c c c c c c c c c c c}
    \toprule
    & \multicolumn{5}{c}{LIBERO-PRO, L10} & \multicolumn{6}{c}{LIBERO-PRO, mean over $4$ suites} \\
    \cmidrule(lr){2-6}\cmidrule(lr){7-12}
    Auxiliary target & Lang. & Pos. & Obj. & Task & Env. & Lang. & Pos. & Obj. & Task & Env. & All \\
    \midrule
    None ($\pi_{0.5}$ baseline)    & $65$ & $8$  & $79$ & $15$ & $47$ & $86.5$ & $46.8$ & $91.0$ & $44.5$ & $74.8$ & $68.7$ \\
    DIFF $+$ ID target             & $63$ & $24$ & $85$ & $16$ & $55$ & $87.2$ & $47.5$ & $92.5$ & $42.0$ & $76.0$ & $69.0$ \\
    DIFF $+$ ID target (policy seed $1$) & $57$ & $18$ & $69$ & $12$ & $52$ & $84.2$ & $44.2$ & $87.2$ & $39.2$ & $70.5$ & $65.1$ \\
    \quad batch-shuffled target    & $65$ & $15$ & $81$ & $15$ & $60$ & $86.5$ & $42.2$ & $87.8$ & $40.8$ & $72.0$ & $65.8$ \\
    \quad random-init encoder      & $72$ & $14$ & $82$ & $25$ & $48$ & $89.2$ & $43.8$ & $92.5$ & $40.0$ & $72.5$ & $67.6$ \\
    V-JEPA 2 $+$ ID (w $0.01$)     & $54$ & $23$ & $79$ & $19$ & $55$ & $84.2$ & $45.5$ & $90.8$ & $41.0$ & $75.8$ & $67.5$ \\
    V-JEPA 2 $+$ ID (w $0.05$)     & $71$ & $8$  & $80$ & $18$ & $47$ & $86.5$ & $40.5$ & $88.8$ & $40.5$ & $72.8$ & $65.8$ \\
    Web-DINO                       & $56$ & $14$ & $83$ & $16$ & $54$ & $83.5$ & $39.8$ & $92.0$ & $35.2$ & $74.5$ & $65.0$ \\
    \bottomrule
    \end{tabular}
    \caption{\textbf{Full LIBERO-PRO results} (success rate, \%), including per-suite shift means. In-distribution results are in Table~\ref{tab:closedloop}; for the omitted V-JEPA $2$ $+$ ID weight-$0.05$ run, Spatial/Object/Goal/L10 are $97/99/82/86$ (mean $91.0$). The second DIFF $+$ ID-target policy seed was evaluated on LIBERO-PRO only.}
    \label{tab:closedloop_full}
\end{table}

\end{document}